\documentclass[10pt,twocolumn,letterpaper]{article}

\usepackage[pagenumbers]{cvpr} 

\usepackage{algorithm}
\usepackage{algpseudocode}

\usepackage[dvipsnames]{xcolor}

\definecolor{cvprblue}{rgb}{0.21,0.49,0.74}
\usepackage[breaklinks,colorlinks,citecolor=cvprblue]{hyperref}

\def\paperID{10468} 
\def\confName{CVPR}
\def\confYear{2024}

\title{Zero-TPrune: Zero-Shot Token Pruning through Leveraging of the Attention Graph in Pre-Trained Transformers}

\author{Hongjie Wang, Bhishma Dedhia, Niraj K. Jha \\
Princeton University \\
Princeton, NJ 08540, USA\\
{\small \{hongjiewang, bdedhia, jha\}@princeton.edu}
}

\begin{document}
\maketitle
\begin{abstract}
Deployment of Transformer models on edge devices is becoming increasingly challenging due to the exponentially growing 
inference cost that scales quadratically with the number of tokens in the input sequence. 
Token pruning is an emerging solution to address this challenge due to its ease of deployment on various Transformer 
backbones. However, most token pruning methods require computationally expensive fine-tuning, which is undesirable in many 
edge deployment cases. 
In this work, we propose Zero-TPrune, the first zero-shot method that considers both the importance and similarity of tokens 
in performing token pruning. It leverages the attention graph of pre-trained Transformer models to produce an importance 
distribution for tokens via our proposed Weighted Page Rank (WPR) algorithm.
This distribution further guides token partitioning for efficient similarity-based pruning. Due to the elimination of the 
fine-tuning overhead, Zero-TPrune can prune large models at negligible computational cost, switch between different pruning configurations at no computational cost, and perform hyperparameter tuning 
efficiently. 
We evaluate the performance of Zero-TPrune on vision tasks by applying it to various vision Transformer backbones and 
testing them on ImageNet. Without any fine-tuning, Zero-TPrune reduces the FLOPs cost of DeiT-S by 34.7\% and improves 
its throughput by 45.3\% with only 0.4\% accuracy loss. Compared with state-of-the-art pruning methods that require 
fine-tuning, Zero-TPrune not only eliminates the need for fine-tuning after pruning but also does so with only 0.1\% 
accuracy loss. Compared with state-of-the-art fine-tuning-free pruning methods, Zero-TPrune reduces accuracy loss by up 
to 49\% with similar FLOPs budgets. 
Project webpage: \href{https://jha-lab.github.io/zerotprune/}{https://jha-lab.github.io/zerotprune}.
\end{abstract}
    
\vspace{-1em}
\section{Introduction}
\label{sec:intro}

The Transformer \cite{Transformer} architecture has emerged as a \textit{de facto} workhorse of contemporary machine learning 
paradigms, showing impressive generalization across a swath of tasks including Computer Vision (CV) \cite{alexnet}, 
natural language processing (NLP) \cite{bert}, robotics \cite{robotics}, and games \cite{RL}. At the heart of the architecture 
lies the multi-headed self-attention that dynamically aggregates parallel-processed tokens, yielding a highly 
effective general-purpose computing framework. The implications are particularly apparent in the case of CV where the Transformer's ability to assimilate rich abstractions from large-scale data facilitates strong transfer to downstream tasks, 
outperforming state-of-the-art Convolutional Neural Networks (CNNs) \cite{vit}.

\begin{figure*}[h]
\centering
\includegraphics[width=\linewidth]{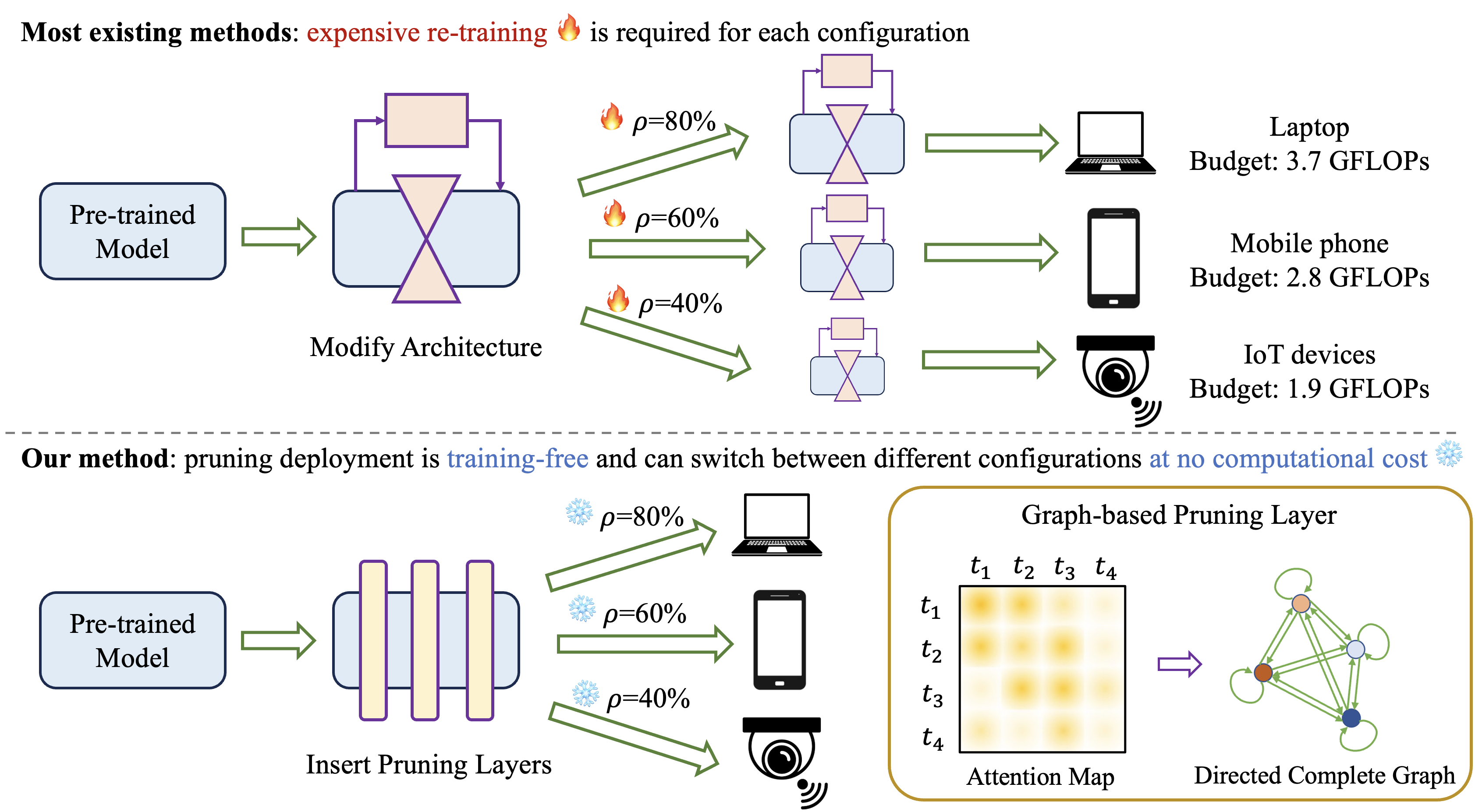}
\vspace{-2em}
\caption{Comparing existing efficiency enhancement methods and Zero-TPrune. $\rho$ represents the retention ratio measured by FLOPS cost. Most existing methods require re-training 
of the model after deploying it; each different pruning configuration requires separate re-training of the model, which is 
extremely expensive. On the contrary, Zero-TPrune is training-free and can 
switch between different pruning configurations 
at no computational cost. This benefits from our graph-based algorithm exploiting correlations between image tokens.}
\vspace{-1.5em}
\label{fig:motivation}
\end{figure*} 

Studies on empirical scaling laws for Vision Transformers (ViTs) \cite{Zhai_2022_CVPR} point to the possibility of 
improvement in model performance with model capacity; recent models have indeed been scaled to billions of parameters.  
While model scaling brings with it the promise of remarkable generalization, it poses serious obstacles to deploying 
such architectures on compute-constrained devices like the edge and executing real-time inference workloads under limited 
energy and memory. \textit{How does one reduce the computational complexity of the forward pass while still maintaining 
the richness of learned representations?} To this end, \textit{Token Pruning} opens up a promising 
avenue.  Drawing a simple analogy to the human vision system, when trying to identify an exotic bird perched on the 
window on an idyllic afternoon, we tend to prune away inconsequential visual details like the cup of piping hot tea lying 
nearby, the ambling pedestrians on the walkway or the sunlit foliage in the background. The attention heads induce quadratic 
computational complexity with respect to the input sequence length. Thus, pruning unimportant tokens can result in 
significant speedups, especially in the case of longer sequences. Since token pruning only prunes the information that 
passes through the sequential layers of the Transformer and does not necessitate architectural modifications to the 
backbone, it can be widely deployed on most Transformer backbones and any computational hardware can fully exploit the 
resultant sparsity. However, most existing token pruning methods rely on token scoring modules that must be trained 
together with the backbone, requiring computationally-expensive re-training or fine-tuning for deployment.
This is impractical for edge applications and users, given the scarcity of computing resources. For example, 
the state-of-the-art token pruning method DynamicViT \cite{dynamicvit} requires 150 hours of fine-tuning on an NVIDIA A100 
GPU to prune the DeiT-S \cite{deit} model. Moreover, the memory and computational resources available may differ widely 
across edge devices; they may also have wide variations in throughput requirements. Fine-tuning-required pruning methods 
need to train the model repeatedly under different pruning configurations imposed by hardware constraints, as shown in 
Fig.~\ref{fig:motivation}, making the pruning process even more expensive. In addition, these methods are impractical for 
pruning very large models due to the high computation overhead of training after pruning. For instance, thousands of A100 GPU 
hours are needed to apply DynamicViT \cite{dynamicvit} to the DeiT-B and DeiT-L models \cite{deit}. 

In this work, we propose a \textit{training-free} zero-shot token pruning method called Zero-TPrune. \textit{How does one 
prune tokens without fine-tuning?} The soft attention between tokens induces a directed graph with tokens as nodes and 
attention as edges. Edge strengths indicate attention values. We posit and later show through a rigorous and comprehensive 
set of experiments that the attention graph is a rich information source for inferring important tokens and, conversely, 
tokens that can readily be pruned. \textit{How does one identify important tokens from the attention graph?} The weights of 
the directed edges on the attention graph can be interpreted as information routing volume between nodes. Utilizing the 
underlying assumption that other important tokens attend to important tokens, we iteratively assign relative importance to 
tokens. Such ranking methods \cite{pagerank} have been ubiquitously used by search engines to organize web pages on the 
Internet. \textit{Can one further exploit redundancy among tokens?} Our experiments show that tokens often learn similar 
abstractions and, therefore, copies of the same feature can be pruned without loss of information. We augment 
importance ranking with similarity-driven pruning to account for similar tokens. Although Zero-TPrune can potentially be 
applied to any Transformer-based tasks, we focus on vision tasks to evaluate its performance in this article. 

The main contributions of this work can be summarized as follows. (1) We present Zero-TPrune, a zero-shot token pruning 
method that efficiently leverages the feature identification capability (considers the attention matrix as an adjacency 
matrix of a directed graph) of pre-trained Transformers. It exploits both the importance and similarity of tokens to perform 
pruning.  (2) We use a graph signal to represent the importance score distribution on tokens and propose the Weighted Page 
Rank (WPR) algorithm to infer unimportant tokens during iterative importance assignment. This iterative scheme reduces 
noise from unimportant tokens during assignment.  (3) Instructed by the importance distribution, we partition tokens into 
two groups and perform similarity-based pruning. Input-dependent partitioning controls the importance distribution of 
tokens pruned by the similarity metric.  (4) We apply Zero-TPrune and baseline methods to various Transformer backbones 
and evaluate their performance on ImageNet \cite{imagenet}. The used backbones include DeiT \cite{deit}, LV-ViT \cite{lvvit}, 
AugReg \cite{augreg}, etc. Compared with state-of-the-art fine-tuning-required Transformer pruning methods, Zero-TPrune 
eliminates the need for fine-tuning after pruning DeiT-S with only around 0.1\% accuracy reduction while achieving the same 
FLOPs saving. Moreover, Zero-TPrune outperforms state-of-the-art fine-tuning-free methods in terms of both accuracy and 
throughput. 
Zero-TPrune reduces accuracy loss by 33\% on DeiT-S when compared with state-of-the-art fine-tuning-free methods. In terms 
of throughput, Zero-TPrune provides 45.3\% off-the-shelf speed-up at a cost of only 0.4\% in accuracy.


\section{Related Works}
\label{sec:related}


In the first few years after the Transformer model was proposed in 2017, it was mainly employed in the NLP field \cite{bert}. ViT \cite{vit} was the first work to directly apply an \emph{encoder-only} Transformer architecture to non-overlapping image patches in an image classification task without employing any convolution operations. Relative to state-of-the-art CNNs, ViT was able to achieve better performance through large-scale pre-training. DeiT \cite{deit} was another convolution-free Transformer that was trained only on ImageNet \cite{imagenet} and achieved better performance than ViT by relying on several training techniques. Both ViT and its follow-up architectures split the input image into multiple non-overlapping image patches and transform them into tokens for further processing. This provides a new dimension to sparsity exploitation that is quite different from sparsity-enhancing techniques employed in CNNs. 


Most of the previous token pruning works focus on NLP tasks, including PoWER-BERT \cite{powerbert}, Length-Adaptive Transformer \cite{lat}, SpAtten \cite{spatten}, TR-BERT \cite{trbert}, and Learned Token Pruning \cite{learned}. For CV tasks, a typical token pruning work is DynamicViT \cite{dynamicvit}.  
It inserts prediction modules between transformer blocks to predict and drop less informative tokens. The prediction
modules are neural networks that can be jointly trained with the vision Transformer backbone. Instead of using a
deterministic strategy to prune tokens, A-ViT \cite{avit} introduces a stochastic pruning process. It uses adaptive
halting modules to compute the halting probability per token. A token is pruned (i.e., discarded) upon reaching the
halting condition. As a result, the number of tokens is gradually reduced, leading to faster inference. Other recent
works on token pruning for ViT include SPViT \cite{spvit}, TPS \cite{tps}, Adaptive Sparse ViT \cite{savit}, DToP
\cite{dtop}, and HeatViT \cite{heatvit}. Although the pruning methods mentioned above require few or even no extra
parameters for pruning, they require computationally-expensive fine-tuning after pruning. On the contrary, our proposed Zero-TPrune can eliminate the training process after pruning with only 0.1\% accuracy reduction.


There are a few previous works that have explored pruning tokens without requiring fine-tuning. ATS \cite{ats} uses an
inverse transform to accomplish adaptive token sampling based on the importance score distribution. When the importance
scores are concentrated on several tokens, the number of sampled tokens automatically reduces. However, ATS only uses
the attention probability of the classification (CLS) token in the attention matrix and ignores the effect of
similarity between tokens. ToMe \cite{tome}, on the other hand, focuses on merging tokens instead of pruning them,
thereby reducing the inference overhead of pre-trained Transformers without fine-tuning. Tokens are progressively
merged based on their similarity as the layers become deeper. However, ToMe solely relies on embedding vectors from
pre-trained models and its matching process lacks appropriate guidance (more details in Section \ref{s-stage}). In contrast, 
Zero-TPrune effectively utilizes both the complete attention matrix and embedding vectors from pre-trained Transformers, 
while simultaneously considering the importance and similarity of tokens. 

\section{Methodology}
\label{sec:methodology}


\begin{figure*}[h]
\centering
\includegraphics[width=\linewidth]{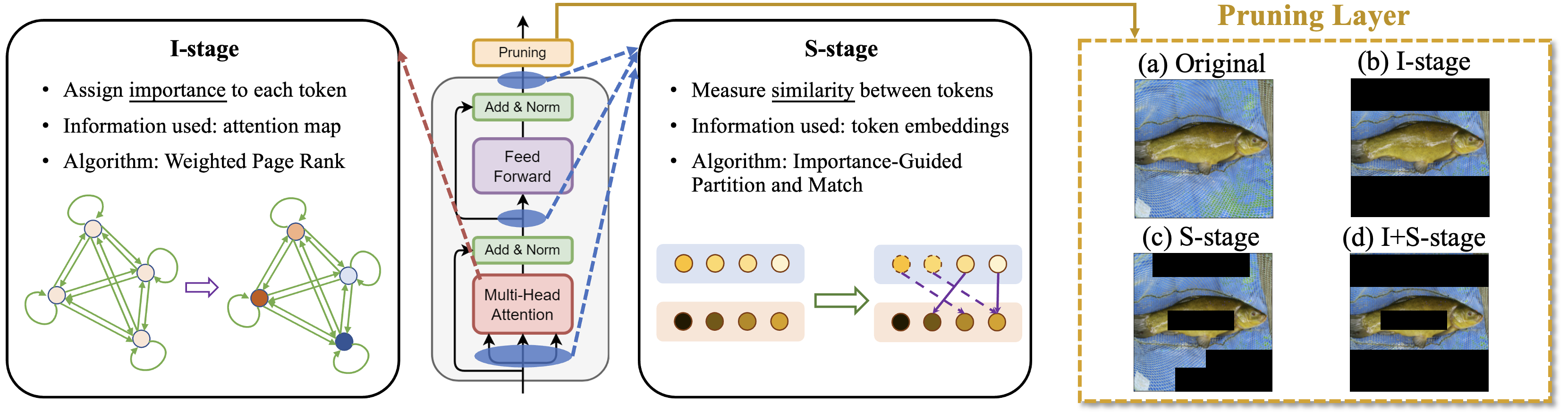}
\vspace{-2em}
\caption{The overall Zero-TPrune framework. Pruning layers can be inserted between Transformer blocks to reduce 
the number of tokens. Pruning layers comprise \textbf{I-stage} and \textbf{S-stage}: \textbf{I-stage} aims at pruning 
unimportant tokens of an image, such as background tokens (see (b)); \textbf{S-stage} aims at pruning tokens that are too 
similar to others, such as repetitive texture tokens (see (c)). A combination of the stages then maximally exploits token 
redundancy (see (d)).}
\vspace{-1.5em}
\label{fig:overview}
\end{figure*} 

In this section, we first provide an overview of Zero-TPrune in
Section~\ref{overview}, then describe its components, \textbf{I-stage} (Section \ref{pagerank}) and \textbf{S-stage} (Section \ref{s-stage}). 
Note that Zero-TPrune is potentially differentiable, which enables the pruned model to be further fine-tuned for better performance. This optional training-after-pruning paradigm is described in Supplementary Material Section \ref{app:fine-tune}.


\subsection{Overview: Zero-TPrune}
\label{overview}


The overall Zero-TPrune framework is shown in Fig.~\ref{fig:overview}. Each pruning layer is composed of multiple stages and can be inserted anywhere between Transformer blocks. The \textbf{I-stage} and \textbf{S-stage} enable Zero-TPrune to take both importance and similarity into consideration. The objective of the \textbf{I-stage} is to obtain an importance score distribution on tokens and retain the top-$k$ important tokens. To achieve this objective, we propose the WPR algorithm and use the attention matrix from the pre-trained Transformer block. In the \textbf{S-stage}, we measure the similarity between tokens based on their embedding vectors and retain only one token in the top-$r$ similar pairs. To reduce computational overheads from all pair-wise combinations, we partition tokens into bipartite groups. Tokens in the same group are never paired to measure similarity. To have improved control over the importance distribution of pruned tokens, we guide the partitioning by their importance rank. 



A straightforward way to combine the two stages is to consecutively connect the \textbf{I-stage} and \textbf{S-stage}: some tokens are pruned based on the obtained importance score distribution in the \textbf{I-stage}; this distribution is then used to guide the partition in the \textbf{S-stage} and some other tokens are pruned based on similarity. However, we empirically observed that such a trivial combination may cause the semantically unimportant tokens to eventually crowd out semantically significant tokens in the \textbf{I-stage}. For example, sometimes background tokens received high importance scores compared to the main object tokens. Details of this phenomenon can be found in Supplementary Material Section \ref{app:overwhelm}. We resolve this issue by interchanging the \textbf{I-stage} and \textbf{S-stage}. This method enables the early elimination of similar tokens in the \textbf{S-stage}, consequently reducing the adverse impact of similarity in the \textbf{I-stage} to a significant extent. We present a comparison of the two patterns in Supplementary Material Section \ref{app:comparison}. To facilitate partitioning in the \textbf{S-stage}, we introduce a pre-ranking \textbf{I$^\prime$-stage} to assign importance scores to tokens with a \emph{single} round of voting. Notably, no token is pruned in the \textbf{I$^\prime$-stage}. Consequently, the pruning layer comprises sequential application of \textbf{I$^\prime$-stage}, \textbf{S-stage}, and \textbf{I-stage}.



\subsection{I-stage: Importance-based Pruning}
\label{pagerank}

\begin{figure}[h]
\centering
\includegraphics[width=\linewidth]{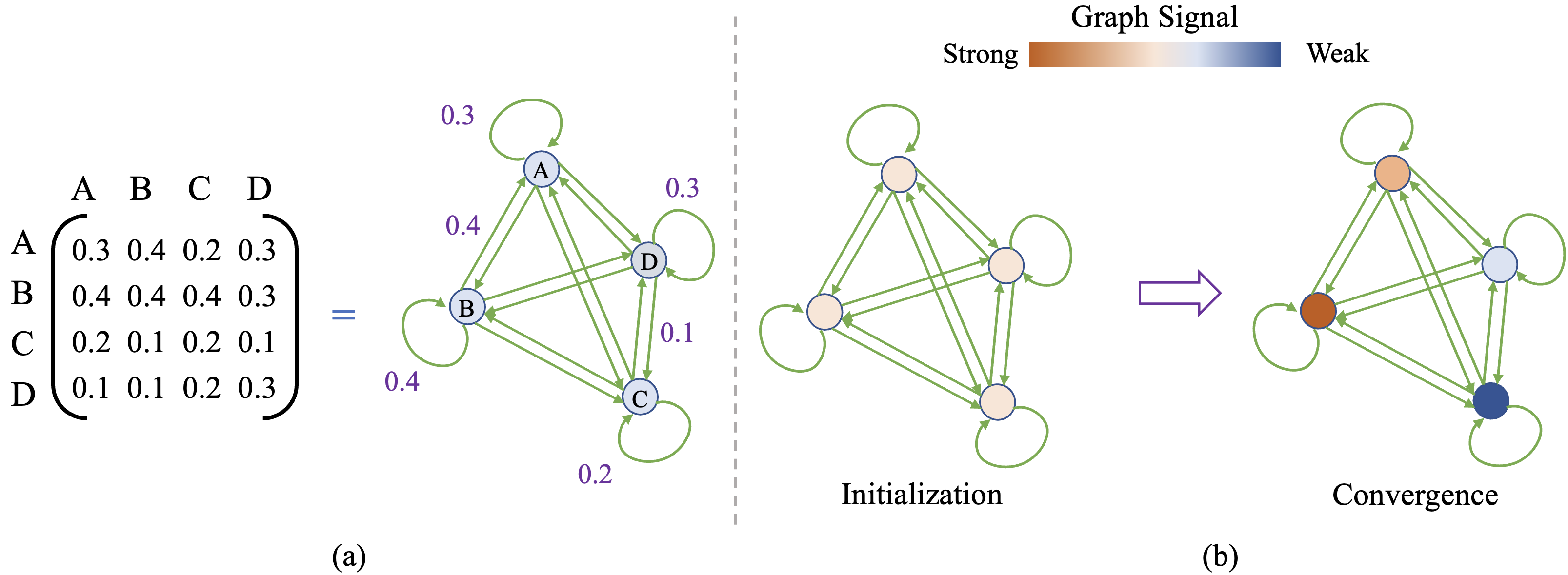}
\vspace{-2em}
\caption{Overview of the I-stage: (a) from a 4$\times$4 attention matrix to an attention graph and (b) graph signal transformation from initialization to convergence.}
\vspace{-1.5em}
\label{fig:attngraph}
\end{figure}

To retain the top-$k$ important tokens, we introduce a ranking metric called \emph{importance score}. In order to obtain the importance score, we treat attention matrix $\mathbf{A}^{(h,l)}$ as the adjacency matrix of a complete, directed graph, called the \emph{attention graph}, as shown in Fig.~\ref{fig:attngraph}(a). Ranking nodes in the graph is challenging because of several reasons. (i) The attention graph is dense, usually including hundreds of nodes and many more edges when the input is an image. (ii) We have a strict budget for the computational overhead incurred by the per-image algorithm during inference. 



Inspired by the Page Rank \cite{pagerank} algorithm, we propose a WPR algorithm to derive the importance scores. Page Rank was used in Google Search for ranking web pages. In the original Page Rank algorithm, links between web pages are unweighted. In order to apply it to the weighted and directed attention graph, we consider the signal of each node in this graph as the importance of each token. We initialize the graph signal uniformly and use the adjacency matrix as a Graph Shifting Operator (GSO). When the GSO is applied to the graph signal, each node votes for which node is more important through the weight assigned to output edges, i.e., the attention that a token pays to other tokens. 
If the node itself is more important, the voting of this node is more important. This is shown in Algorithm \ref{alg:prgraph}. The transition from initialization to convergence is shown in Fig.~\ref{fig:attngraph}(b).


\begin{algorithm}
\caption{Graph-based Weighted Page Rank (WPR) algorithm}\label{alg:prgraph}
\begin{algorithmic}
\Require $N>0$ is the number of nodes in the graph; $A\in \mathbb{R}^{N \times N}$ is the adjacency matrix of this graph; $s\in \mathbb{R}^{N}$ represents the graph signal
\Ensure  $s\in \mathbb{R}^{N}$ represents the importance score of nodes in the graph
\State $s^{0} \gets \frac{1}{N} \times \textbf{$e_N$}$ \Comment{Initialize the graph signal uniformly}
\State $t \gets 0$
\While{$(|s^{t} - s^{t-1}|> \epsilon) \textbf{or} (t=0)$} \Comment{Continue iterating if not converged}
    \State $t \gets t+1$
    \State $s^{t} \gets A^T \times s^{t-1}$  \Comment{Use the adjacency matrix as a graph shift operator}
\EndWhile
\State $s \gets s^{t}$
\end{algorithmic}
\end{algorithm}

We obtain the expression for the importance score of each node (i.e., token) in the $l$-th layer, $h$-th head as follows:

\vspace{-2em}

\begin{equation}
s^{(h, l)}\left(x_i\right)= \frac{1}{N} \sum_{j=1}^{N} \mathbf{A}^{(h, l)}\left(x_i, x_j\right) \cdot s^{(h, l)}\left(x_j\right)
\end{equation}


\vspace{-1em}
\noindent
where $s^{(h, l)} (x_j)$ is the importance score of node $x_i$ in the $h$-th head of the $l$-th layer, and $N$ is the number of tokens in the $l$-th layer . $s^{(h,l)}(x_i)$ is derived from the weighted sum of the received attention. 
WPR thus recursively assigns high importance to semantically significant tokens and reduces noise from unimportant semantically weak tokens.
We retain the top-$k$ important tokens ($k$ is determined by the retention rates and the total number of tokens).



Simply averaging the importance scores across different heads is not the optimal choice. Different heads in an encoder layer usually pay attention to different parts of the input image (a visual example is given in Supplementery Material Section \ref{app:eip}). Thus, there are some tokens that are very important in one or two heads, but not in others. On the other hand, some tokens have low-to-medium importance in all heads. The former tokens are usually more informative than the latter tokens. However, if there are multiple heads and the importance score is directly averaged across all heads, the latter tokens may get the same or even higher score than the former tokens, leading to incorrect token ranking and pruning. In order to address this problem, we aggregate the importance scores across heads via a root-mean of sum of squares. We call this the Emphasizing Informative Region (EIR) aggregation. 
We observe that EIR effectively distinguishes informative areas from non-informative ones. A concrete example that compares EIR with other methods (such as \textit{argmax} and average) is given in Supplementary Material Section \ref{app:eip}.

Besides the issue mentioned above, sometimes the importance scores given by the WPR algorithm may converge to an undesired distribution in some heads: (1) tokens at the edge of the input image may get very high importance scores; (2) the importance score distribution may become nearly uniform. We provide visual examples of these cases in Supplementary Material Section \ref{app:vhf}. Both heads in these cases do not provide helpful information and are even misleading. To mitigate the negative impact of these heads, we introduce the Variance-based Head Filter (VHF). We compute the variance of the distribution in each head and set both a minimum and a maximum threshold for the variance. Heads with a distribution variance exceeding the maximum threshold or falling below the minimum threshold are excluded from the computation. Then the final importance score equation becomes: 

\vspace{-1.5em}

\begin{equation}
s^{(l)}\left(x_{i}\right)=\sqrt{\frac{\sum_{h=1}^{N_h} {s^{(h,l)}(x_i)}^2 \cdot \eta \left(v_{\text{min}} \leq Var_h \leq v_{\text{max}}\right)}{\sum_{h=1}^{N_h} \eta \left(v_{\text{min}} \leq Var_h \leq v_{\text{max}}\right)} 
} 
\end{equation}


\noindent
where $\eta \left(v_{\text{min}} \leq Var_h \leq v_{\text{max}}\right)$ equals 1 if $v_{\text{min}} \leq Var_h \leq v_{\text{max}}$, otherwise it equals 0; $v_{\text{min}}$ and $v_{\text{max}}$ represent the minimum and maximum threshold, respectively; $Var_h$ is the importance score variance of tokens in the $h$-th head; $N_h$ is the number of heads in the $l$-th layer. The complexity of \textbf{I-stage}, including WPR, EIR, and VHF, is $O(N^2)$, where $N$ is the number of tokens.


\subsection{S-stage: Similarity-based Pruning}
\label{s-stage}







\begin{figure}[ht]
\centering
\includegraphics[width=\linewidth]{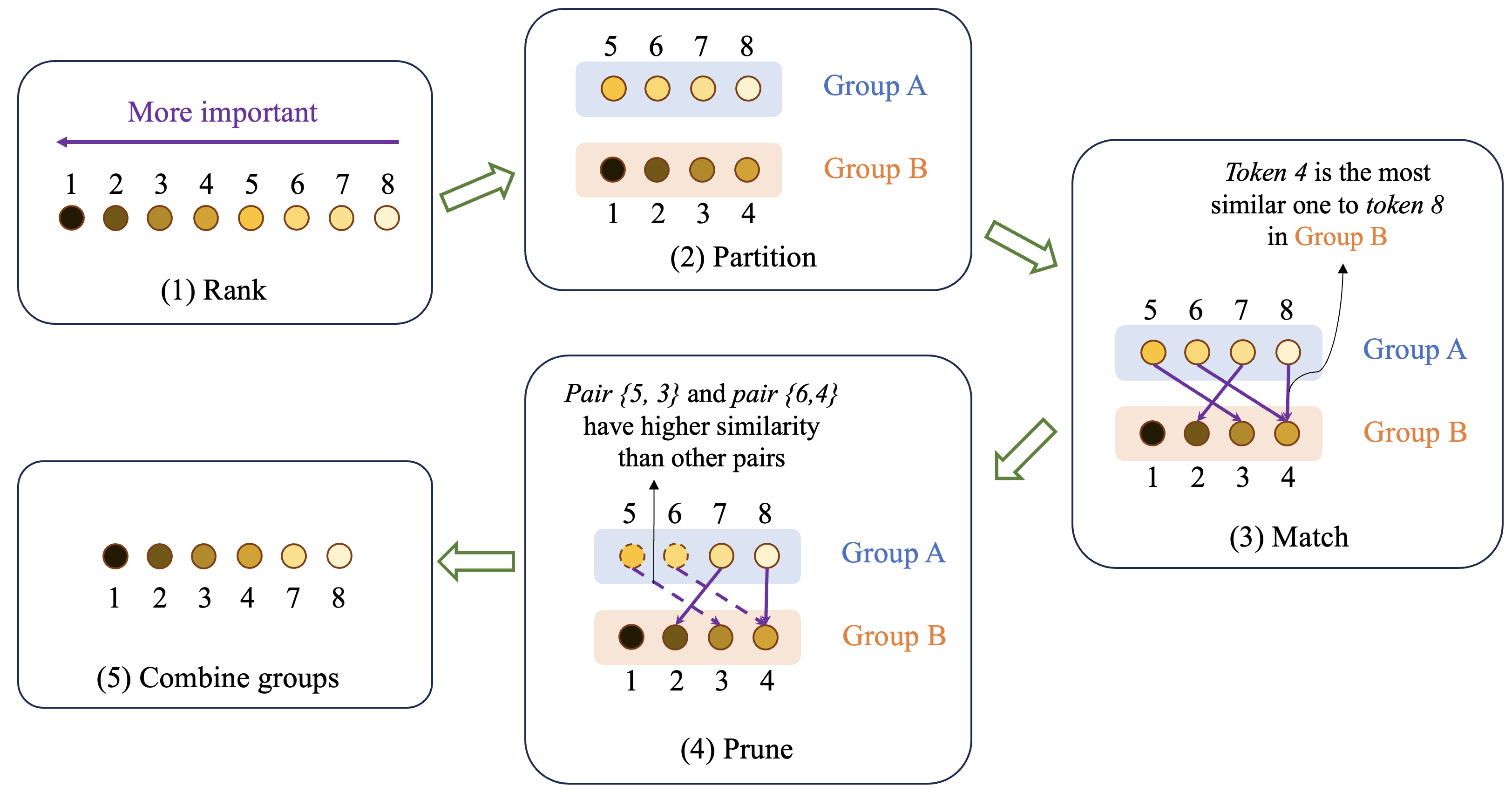}
\vspace{-2em}
\caption{The importance-based pruning process in the S-stage. As an example, sequential partitioning (pruning 
unimportant part) is used in this figure.}
\label{fig:s-stage}
\end{figure}

As discussed previously, it is valuable to measure similarity even between important tokens and perform further pruning. 
Previous work \cite{tome} uses image-agnostic token partitioning to measure pair-wise similarity. Instead, we
propose a per-image importance-driven partition for similarity pruning, as shown in Fig.~\ref{fig:s-stage}. 



\vspace{-1em}

\vspace{-1em}

\textbf{Fig.~\ref{fig:s-stage} (1\&2):} Based on the importance score of tokens, we sequentially partition them
into groups of roughly equal size, $A$ and $B$, and prune the less important group. We explore other
importance-guided partitioning schemes, including alternative partitioning and random partitioning, and provide
ablation results in Supplementary Material Section \ref{app:designchoice}. \textbf{Fig.~\ref{fig:s-stage} (3):} We
then identify the most similar token in Group $B$ for each token in Group $A$ and record the corresponding
similarity of each pair. To accomplish this, we represent each token by a feature vector, which can be derived
from several available choices, such as corresponding vectors in the Key, Query, or Value matrix. Our ablation
experiments indicate that using vectors from the Key matrix is the optimal choice. We compute similarity on these
vectors using a designated metric, such as cosine similarity, Manhattan distance, or Euclidean distance. Following
the results of our ablation experiments, we employ cosine similarity. We provide a detailed account of our
ablation experiment outcomes in Supplementary Material Section \ref{app:designchoice}.
\textbf{Fig.~\ref{fig:s-stage} (4\&5):} In the next step, we select top-$r$ similar pairs and prune corresponding
tokens in Group $A$. We prune one token in each selected pair instead of merging them, due to the following
reasons: (i) since tokens in the selected pairs are similar, pruning one of them results in minimal
information loss; (ii) merged tokens should have higher weights in the following computation \cite{tome}, which
makes it incompatible with certain backbones, such as Sparse Transformer \cite{sparseattn}. Finally, we pass the 
remaining tokens to the next stage. The complexity of \textbf{S-stage} is $O(N^2\times d)$, where $N$ is the number of tokens and $d$ is the dimension of token embeddings.

Importance-guided partitioning in the \textbf{S-stage} facilitates stable control over the importance of the pruned tokens for different input images. By pruning similar tokens instead of merging them, our method maintains compatibility with certain specialized backbones \cite{sparseattn} while incurring only minimal information loss.

\section{Experimental Results}
\label{sec:experiment}


\begin{figure*}[h]
\centering
\includegraphics[width=\linewidth]{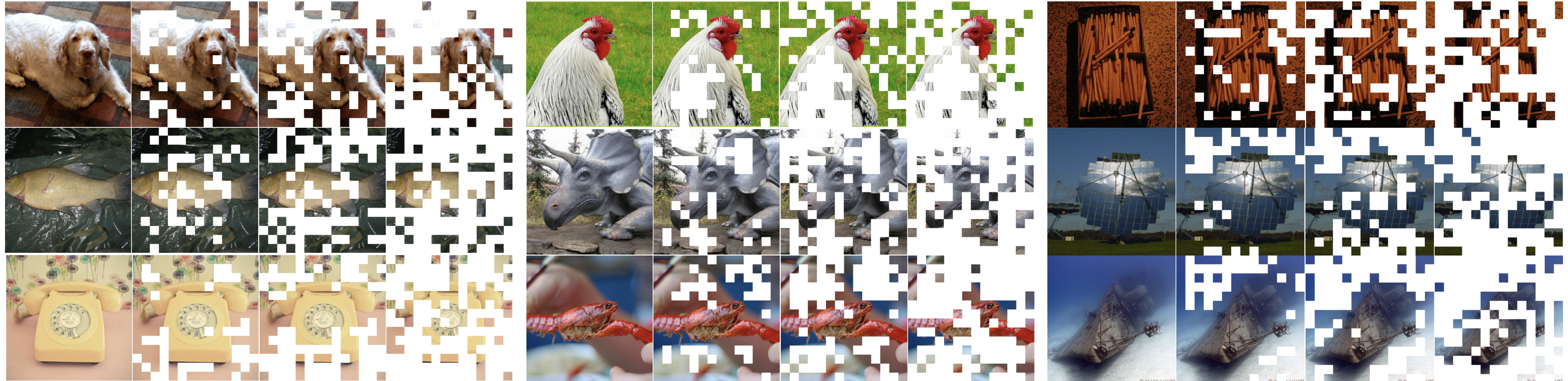}
\vspace{-2em}
\caption{Visualized examples of the pruning process conducted by Zero-TPrune. Images are randomly selected from ImageNet validation dataset. When the pruning rate is aggressive and the main object occupies most of the image area, it is not enough to only prune background tokens. Zero-TPrune exploits similarity between main object tokens and prunes redundant ones.}
\vspace{-1.5em}
\label{fig:visualize_prune}
\end{figure*} 

In this section, we first describe the visualized token pruning process of several images in the ImageNet validation dataset, as shown in Fig. \ref{fig:visualize_prune}. We also present ablation experiments to validate our design choices and the effectiveness of our proposed methods. Then, we compare Zero-TPrune with state-of-the-art token pruning methods. 

\textbf{Experimental Setup:} To compare different pruning methods, we apply them to various vision
Transformer backbones and evaluate the performance of pruned models on ImageNet \cite{imagenet}.
We evaluate the models on 224px images unless otherwise noted. We estimate the inference GFLOPS on a V100 GPU using the \texttt{fvcore}\footnote{\url{https://github.com/facebookresearch/fvcore}} library. We measure the inference throughput of pruned models 
on a single A100 GPU and perform fine-tuning after pruning on A100 GPUs.

Although our experiments focus on the classification task, Zero-TPrune can potentially be applied to other tasks, 
such as generation and segmentation. We refer to the design that can be transferred to other tasks as 
\textbf{``Zero-TPrune-uni"}, meaning ``Zero-TPrune for universal purpose." For the classification task, the CLS token 
is known to be much more important than other tokens, and its attention weights are a much stronger signal for 
selecting tokens \cite{ats}. Thus, instead of initializing the importance score of tokens uniformly, we assign the 
CLS token an importance score that is $\sqrt{N}$ times larger than other tokens during initialization in 
the \textbf{I-stage}, where $N$ is the number of tokens. We call this design \textbf{``Zero-TPrune"} in the following 
experiments. 

For ablation experiments, we implement Zero-TPrune on the DeiT-S model \cite{deit} with different configurations and 
design choices. For comparisons with state-of-the-art token pruning works, we divide them into two types: (1) methods 
that require fine-tuning of the pruned model, including DynamicViT \cite{dynamicvit} and A-ViT \cite{avit}; 
(2) fine-tuning-free methods, including ATS \cite{ats} and ToMe \cite{tome} (which is a token-merging instead of a 
token-pruning method, but is also fine-tuning-free). For the first type, we compare implementations on DeiT models. 
We use the official implementation of DynamicViT
to reproduce its results and also generate some new ones for comparison. For A-ViT, we directly use
the results presented in that article. For the second type, we use the official open-source code of
ATS and ToMe to implement them on various pre-trained Transformer backbones, including DeiT
\cite{deit}, LV-ViT \cite{lvvit}, MAE \cite{mae}, AugReg \cite{augreg}, and SWAG \cite{swag}. We
compare the off-the-shelf performance of Zero-TPrune with theirs. In addition, we compare the performance
of pruned models on downstream tasks to check their transfer learning capability, following the
selection of datasets in \cite{idmm}. We provide details of the selected datasets in Supplementary
Material Section \ref{app:downdata}. Note that ToMe has an optional design, Proportional Attention (PA), that is 
dedicated to classification \cite{tomesd} and is not compatible with sparse attention design \cite{sparseattn}. We 
call ToMe with PA disabled ``ToMe-uni" and ToMe with PA enabled ``ToMe." ATS is a method solely based on the CLS 
token; hence, it does not have a universal version.

To further validate the effectiveness of Zero-TPrune, we supplement comparisons with depth-adaptive methods and other straightforward attention-based token ranking methods (e.g., averaging the received attention) in Supplementary Material Section \ref{app:depthada} and \ref{app:otherattn}.


\subsection{Ablation Experiments}


We use ablation experiments to determine the optimal hyperparameters for Zero-TPrune. First, it is computationally
expensive to check whether the WPR algorithm converges after each iteration. Thus, it would be desirable if we
could determine the number of its iterations in advance. By checking the importance distributions after different
numbers of iterations and computing the Kullback-Liebler (KL) divergence between them, we find 30-50, 5-10, and 1
iteration(s) are enough to ensure convergence in the first three layers, medium layers, and last three layers,
respectively. We provide visual and quantitative comparisons in Supplementary Material Section \ref{app:converge}. Second, we determine good enough minimum and maximum thresholds for VHF through random initialization and greedy search. We provide detailed search configurations and results in Supplementary Material Section \ref{app:range-vhf}. The range found is [0.01,0.7], which is the default setting in our experiments. Third, we explore optimal design choices in the \textbf{S-stage} with ablation experiments presented in Supplementary Material Section \ref{app:designchoice}.

To illustrate the effectiveness of the different techniques employed in Zero-TPrune, we break down their contribution. We apply different combinations of techniques employed in Zero-TPrune to the DeiT-S model and evaluate the performance of the pruned models. 
We insert pruning layers after the [1,3,6,9,11]-th layer with a retention rate of [1,0.9,0.8,0.7,1] and \#iterations of 
[30,5,5,1,1] in the \textbf{I-stage}, and prune 10 tokens in each \textbf{S-stage}. Before adding the 
\textbf{S-stage}, we insert pruning layers after the [3,6,9,11]-th layer with a retention rate of [0.8,0.7,0.7,0.6] 
and \#iterations of [5,5,1,1]. We show the results in Table \ref{tab:ablation} (we provide corresponding results of 
Zero-TPrune-uni in Supplementary Material Section \ref{app:zero-tp-uni}). The WPR algorithm improves 
the performance significantly. The EIR/VHF techniques and the \textbf{S-stage} improve the performance further. 


\begin{table}[htbp]
  \scriptsize
  \centering
  \vspace{-0.5em}
  \caption{Contribution breakdown of the different techniques employed in Zero-TPrune. The used batch size is 512.}
  \vspace{-0.5em}
  {
  \setlength\tabcolsep{3pt}
    \begin{tabular}{lllll}
    \toprule
    Acc@1  & Params & FLOPS/img & Throughput & Method \\
    \midrule
    79.8\% (base)   & 22M   & 4.55G & 1505.9 img/s & Unpruned model \\
    76.8\% (-3.0\%) & 22M   & 3.08G & 2164.4 img/s & random drop \\
    78.6\% (+1.8\%) & 22M   & 3.08G & 2136.5 img/s & WPR \\
    78.8\% (+0.2\%) & 22M   & 3.08G & 2132.6 img/s & WPR+EIR \\
    78.9\% (+0.1\%) & 22M   & 3.08G & 2103.1 img/s & WPR+EIR+VHF (I-stage) \\
    79.4\% (+0.5\%) & 22M   & 3.08G & 2063.9 img/s & I-stage + S-stage \\
    
    \bottomrule
    \end{tabular}%
    }
  \label{tab:ablation}%
  \vspace{-1em}
\end{table}%


To further improve the performance of pruned models, we can employ Monto Carlo Simulation (MCS) to randomly explore the hyperparameter space, which
includes the number and location of pruning layers, corresponding retention rates, number of iterations in each layer, and number of tokens to be pruned in each \textbf{S-stage}. After conducting thousands of trials, we select the optimal setting that exhibits the best performance achieved by Zero-TPrune while maintaining a fixed GFLOPS budget. In the case shown in Table \ref{tab:ablation}, MCS helps to achieve 79.5\% accuracy with 3.08 GFLOPS. To ensure a fair comparison, we do not use MCS in Zero-TPrune in subsequent comparisons with state-of-the-art methods. Zero-TPrune is not sensitive to 
the hyperparameter choice, as illustrated with experimental results in Supplementary Material Section \ref{app:hyper-search}.


\subsection{Comparison with State-of-the-Art Methods}


In this section, we choose the number and location of pruning layers with either constant or
uniformly declining retention rates to match the given GFLOPS budget. We keep the number of pruned tokens in 
each \textbf{S-stage} constant. We fix the number of iterations to 30, 5, and 1 for the first three, 
intermediate, and the last three layers, respectively.


\begin{figure}
    \centering
    \includegraphics[width=\linewidth]{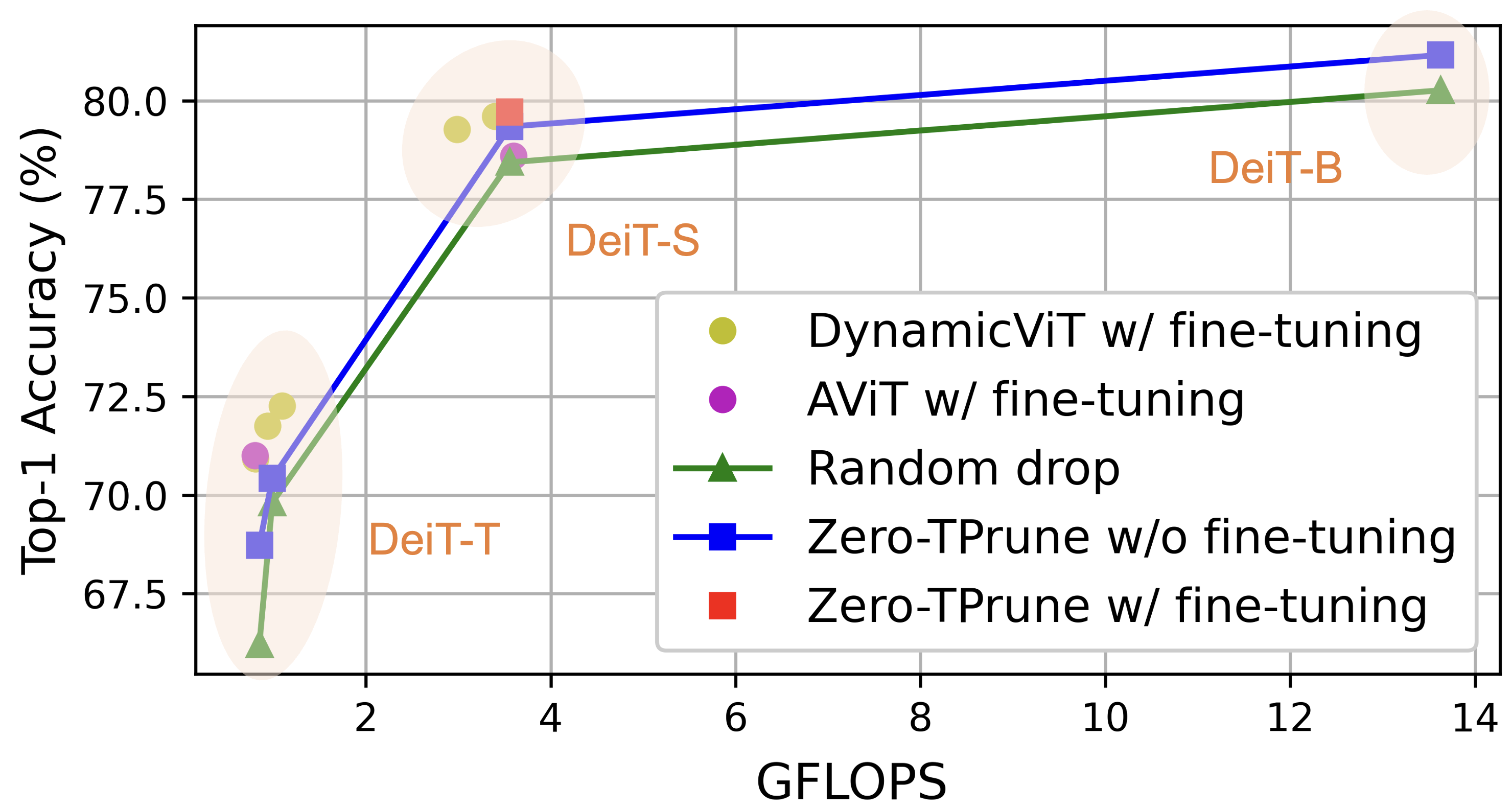}
    \vspace{-2em}
    \caption{Performance comparison between Zero-TPrune and state-of-the-art fine-tuning-required methods.}
    \vspace{-2em}
    \label{fig:acc_flops}
\end{figure}

\noindent\textbf{Comparison with Fine-Tuning-Required Methods:} To illustrate the advantages of Zero-TPrune, we compare its 
performance with that of DynamicViT and A-ViT with/without fine-tuning after pruning. Given the random initialization and 
the fact that the pruning modules in DynamicViT and A-ViT need to be trained, the performance of DynamicViT and A-ViT 
without fine-tuning after pruning is based on randomly-pruned tokens. Fig.~\ref{fig:acc_flops} clearly demonstrates the advantages of Zero-TPrune over state-of-the-art
fine-tuning-required pruning methods, i.e., DynamicViT and A-ViT. Without fine-tuning
after pruning, Zero-TPrune outperforms DynamicViT and A-ViT (using random drop in this case) by around
1\%. This means Zero-TPrune \textbf{reduces the accuracy drop by more than 60\%}. The performance of
Zero-TPrune without fine-tuning after pruning is comparable to that of DynamicViT and A-ViT with
fine-tuning after pruning (e.g., 0.1\% accuracy loss relative to the best, given a 3.5 GFLOPS budget
on DeiT-S). With fine-tuning after pruning, Zero-TPrune outperforms both DynamicViT and A-ViT.
Zero-TPrune can also be easily applied to larger models (e.g., given a 13.6 GFLOPS budget on DeiT-B) for higher accuracy. On the contrary, applying DynamicViT and A-ViT to large models is very computationally expensive due to their expensive fine-tuning after pruning.


\noindent\textbf{Comparison with Fine-Tuning-Free Methods:} ATS and ToMe provide an off-the-shelf option to
prune Transformer models without the requirement of fine-tuning after pruning. We first apply them and
Zero-TPrune to the DeiT-S model to compare off-the-shelf performance after pruning without
fine-tuning. The results are shown in Fig.~\ref{fig:accu_loss} and Table~\ref{tab:offtheshelf}. We provide more results related to throughput in Supplementary Material Section \ref{app:throughput}. As
shown in Fig.~\ref{fig:accu_loss}, compared with state-of-the-art fine-tuning-free methods,
Zero-TPrune \textbf{reduces the accuracy loss by 33\%} on the DeiT-S model with a 3 GFLOPS budget. If
we change the pruning configuration and give a lower budget (e.g., reduce GFLOPS by 45\%), the
accuracy loss introduced by Zero-TPrune is still only 0.7\%. Zero-TPrune can \textbf{reduce GFLOPS by 13\% at nearly no cost}. Note that these results are obtained without fine-tuning.

\begin{figure}[htbp]
    \centering
    \vspace{-1em}
    \includegraphics[width=\linewidth]{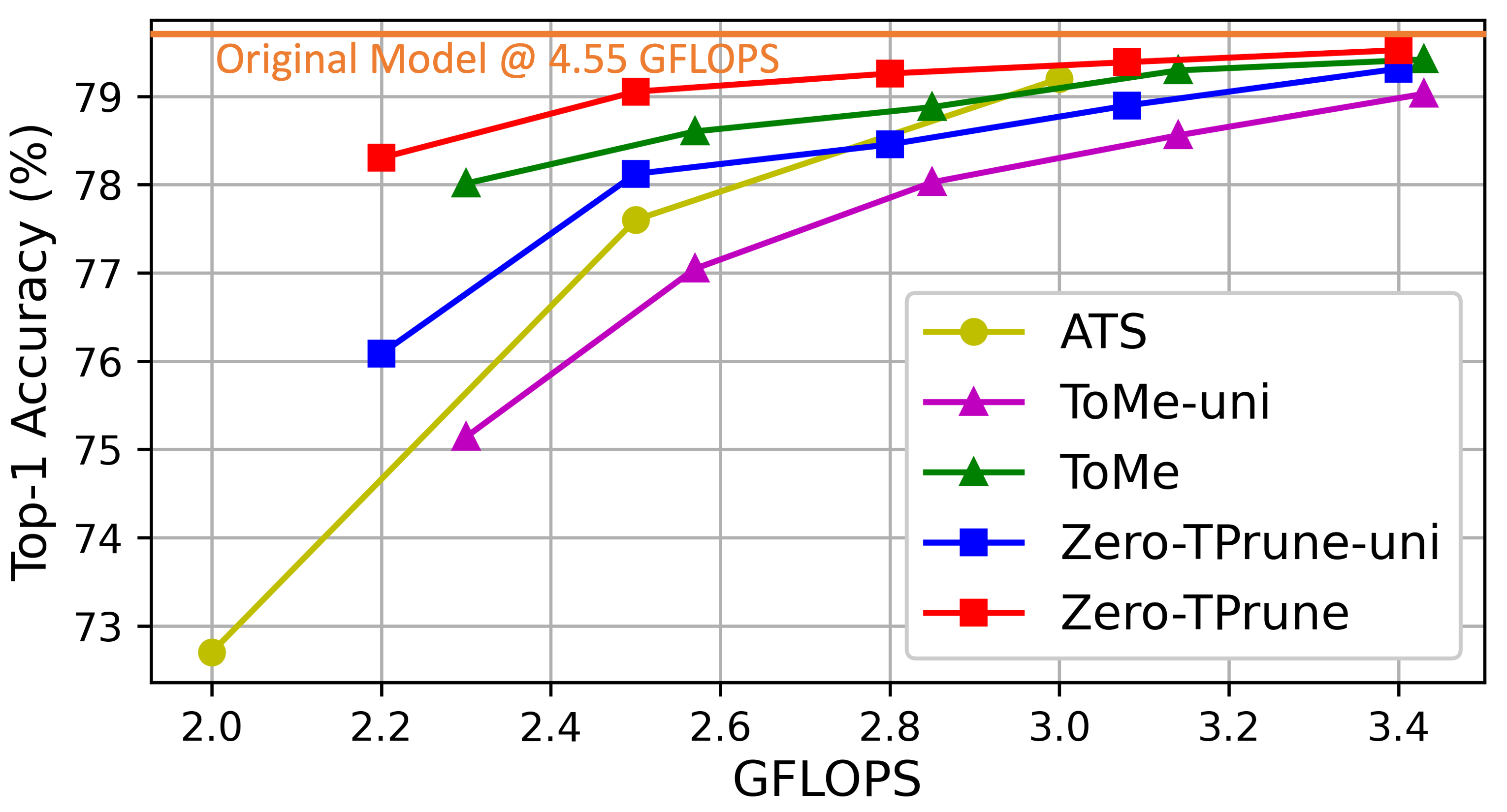}
    \vspace{-2em}
    \caption{Performance comparison between Zero-TPrune and state-of-the-art fine-tuning-free methods. The applied Transformer backbone is DeiT-S.}
    \label{fig:accu_loss}
    \vspace{-1.5em}
\end{figure}


\begin{table}[htbp]
  \scriptsize
  \centering
  \caption{Performance of pruned DeiT-S models without fine-tuning. Throughput is measured on a single NVIDIA A100 GPU.}
  \vspace{-0.5em}
    \begin{tabular}{llll}
    \toprule
    Method  &  Acc@top1    & GFLOPS & Throughput(img/s) \\
    \midrule
    DeiT-S & 79.8\% & 4.55 & 1505.9    \\
    + ATS   & 79.2\% (-0.6\%) & 3.00 (-33.4\%) & 2062.3 (+36.9\%)   \\
    + ToMe & 78.9\% (-0.9\%) & 2.95 (-35.2\%) & 2263.9  (+50.3\%) \\
    + Zero-TP-a   &  \textbf{79.4\% (-0.4\%)}   & 2.97 (-34.7\%) & 2188.4 (+45.3\%) \\
    + Zero-TP-b   &  \textbf{79.1\% (-0.7\%)}   & 2.50 (-45.1\%) & 2458.4 (+63.2\%) \\
    + Zero-TP-c   &  \textbf{79.8\% (-0.0\%)}   & 3.97 (-12.7\%) & 1673.2 (+11.1\%) \\
    \bottomrule
    \end{tabular}%
  \label{tab:offtheshelf}%
  \vspace{-0.5em}
\end{table}%

We further evaluate Zero-TPrune and baselines on various backbones with different sizes. The results
are shown in Table~\ref{tab:more_backbone}. We find that when the
original model is medium-sized, e.g., AugReg and LV-ViT-S, Zero-TPrune outperforms baseline methods by
a large margin (it reduces accuracy loss by up to 49\%). For large models, if the pruning is moderate
(i.e., reduce GFLOPS by 20\%), Zero-TPrune still outperforms baseline methods. However, we found when large models are aggressively pruned (i.e., reduce GFLOPS by 50\%), Zero-TPrune does not outperform baselines. Note that aggressively pruning large models is usually not a good idea, which is indicated by comparing the optimal pruned LV-ViT-M model (\textit{ToMe, 81.6\% with 6.3 GFLOPS}) and the optimal pruned LV-ViT-S model (\textit{Zero-TPrune, 81.5\% with 3.5 GFLOPS}). The latter requires only 60\% of the GFLOPS at the cost of 0.1\% accuracy loss. Compared with aggressively-pruned large models, using a smaller pre-trained model instead is often a better choice. We provide a detailed discussion in Supplementary Material Section \ref{app:scaling}.

We also evaluate the performance of pruned models on downstream tasks to measure their transfer
learning capability. We select several small image datasets for this purpose. Zero-TPrune outperforms baselines on
most datasets, indicating its strong transfer learning capability after pruning. We introduce selected
datasets and present detailed experimental results in Supplementary Material Section \ref{app:downdata}.

\vspace{-0.5em}

\begin{table}[htbp]
  \scriptsize
  \centering
  \caption{Performance of pruned AugReg, LV-ViT, and SWAG models without fine-tuning. SWAG models perform inference on 384px images.}
  \vspace{-0.5em}
  {\setlength\tabcolsep{5pt}
    \begin{tabular}{lll|lll}
    \toprule
    Method  &  Acc@top1    & GFLOPS  & Method  &  Acc@top1    & GFLOPS \\
    \midrule
    AugReg & 81.41\% & 4.55 & MAE & 83.62\% & 55.4   \\
    + ATS & 79.21\%  &  2.80 & +ATS & 82.07\%  &  42.3\\
    + ToMe & 79.30\%  &  2.78 & +ToMe & 82.69\%  &  42.2 \\
    + Zero-TP   & \textbf{80.22\%} & \textbf{2.79}  & +Zero-TP  & \textbf{82.93\%} & 42.3\\
    \midrule
    LV-ViT-S & 83.3\% & 6.6 &    SWAG & 85.30\% & 55.6    \\
    + ATS   & 80.4\% & 3.5  & +ATS & 84.21\%  &  43.8  \\
    + ToMe & 79.8\% & 3.6  & +ToMe & 85.09\%  &  43.8\\
    + Zero-TP   &  \textbf{81.5\% }   & \textbf{3.5} & +Zero-TP   & \textbf{85.17\%} & 43.8 \\
    \bottomrule
    \end{tabular}%
    }
  \label{tab:more_backbone}%
\end{table}%

    
    



    
    


\vspace{-1em}

\section{Conclusion}
\label{sec:conclusion}


In this article, we proposed Zero-TPrune, a zero-shot token pruning method that exploits both the importance and 
similarity of tokens to eliminate the fine-tuning process for pruning. In the \textbf{I-stage}, it considers the 
attention matrix to be an adjacency matrix of an attention graph, which reduces noise from unimportant tokens. In 
the \textbf{S-stage}, it uses importance distribution to guide token partitioning and similarity-based pruning, making 
them more stable and precise.  Through the implementation of Zero-TPrune and baseline methods on various Transformer 
backbones and evaluation on ImageNet, we showed that it can eliminate the fine-tuning process for pruning with
very small accuracy reduction. Moreover, when compared to state-of-the-art off-the-shelf pruning methods, Zero-TPrune 
not only outperforms them by reducing accuracy loss by up to 49\% but also enhances the transfer learning capability of 
pruned models. These findings emphasize the effectiveness of Zero-TPrune in balancing model compression and preservation 
of performance, making it a promising approach for efficient and accurate pruning of Transformer models. Future work can 
enhance the capabilities of Zero-TPrune further. 
One intriguing topic of future study is examining the applicability of Zero-TPrune on tasks such as image 
reconstruction, segmentation, and generation. Investigating the potential benefits and efficiency gains of employing 
Zero-TPrune in these domains holds promise for advancing the field further.

\noindent\textbf{Acknowledgment.} This work was supported by NSF under Grant No. CCF-2203399.

\newpage

{
    \small
    \bibliographystyle{ieeenat_fullname}
    \bibliography{main}
}

\clearpage
\setcounter{page}{1}
\maketitlesupplementary

\appendix
\section{The I-S Pattern and the I$^\prime$-S-I Pattern}

In this section, we first demonstrate the \textit{overwhelming of the major group} issue caused by the I-S pattern and then 
compare it with the I$^\prime$-S-I pattern visually.

\subsection{Overwhelming of the Major Group with the I-S Pattern}
\label{app:overwhelm}

\begin{figure}[h]
\centering
\includegraphics[width=\linewidth]{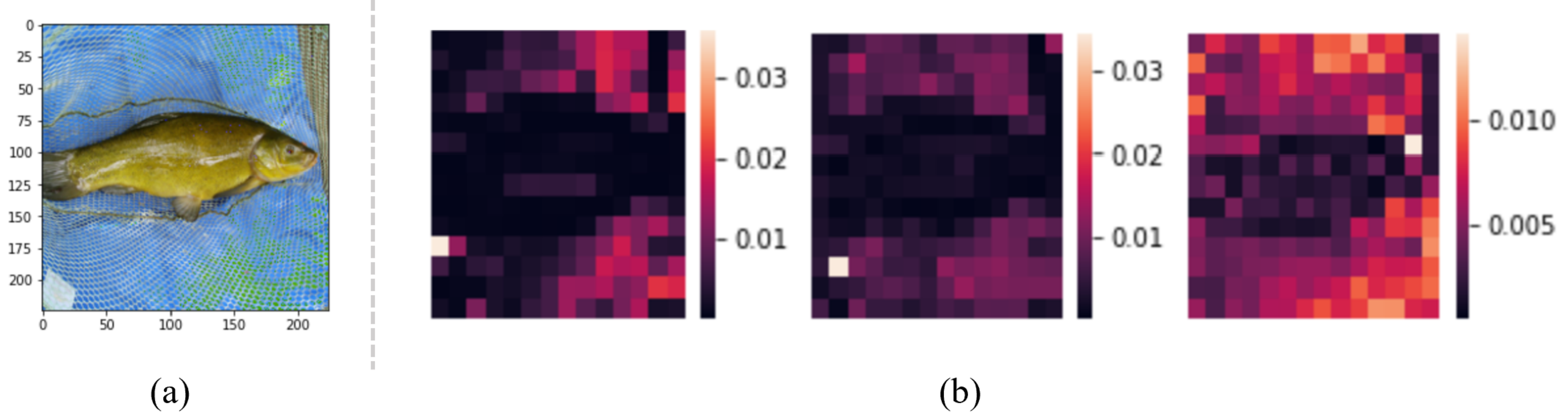}
\caption{An example illustrating that a large unimportant group may overwhelm a small important group:
(a) input image and (b) three examples showing that unimportant background tokens overwhelm the important fish tokens.
}
\vspace{-0.5em}
\label{fig:limit2}
\end{figure}

Sometimes, unimportant parts of an image may be identified as important and important parts as unimportant by our graph-based WPR algorithm. In the \textbf{I-stage}, each token votes for "more important tokens" and the weight of their votes is determined by their importance in the last round of voting. Besides the semantically significant tokens, tokens also intend to vote for tokens \emph{that are similar to them}. When semantically significant tokens (e.g., main object tokens) are only a small part of an image and unimportant background tokens dominate, sometimes background tokens vote for each other and gradually obtain high importance scores after multiple rounds of voting. An example is shown in Fig.~\ref{fig:limit2}.  It shows that the background of the image is considered important in the Transformer heads. The fish itself, surprisingly, is considered unimportant.

In this image, the \textit{background} and \textit{fish} tokens form two sets: \textit{A} and \textit{B}. In the 
beginning, because tokens in \textit{set B} are more semantically significant than those in \textit{set A}, they have relatively high 
importance scores. However, both tokens in \textit{set A} and \textit{set B} mainly intend to vote for tokens in their 
own set. Thus, it is easier for \textit{set A} to form tokens with high importance scores because \textit{set A} includes 
more tokens. These “highly important” tokens have larger voting weights in the next iteration. This makes it even easier 
for other tokens in \textit{set A} to get “high importance.” This is a positive feedback loop, with the result that the 
most “important” tokens end up in \textit{set A}.

\subsection{Comparison}
\label{app:comparison}

\begin{figure*}[h]
\centering
\includegraphics[width=\linewidth]{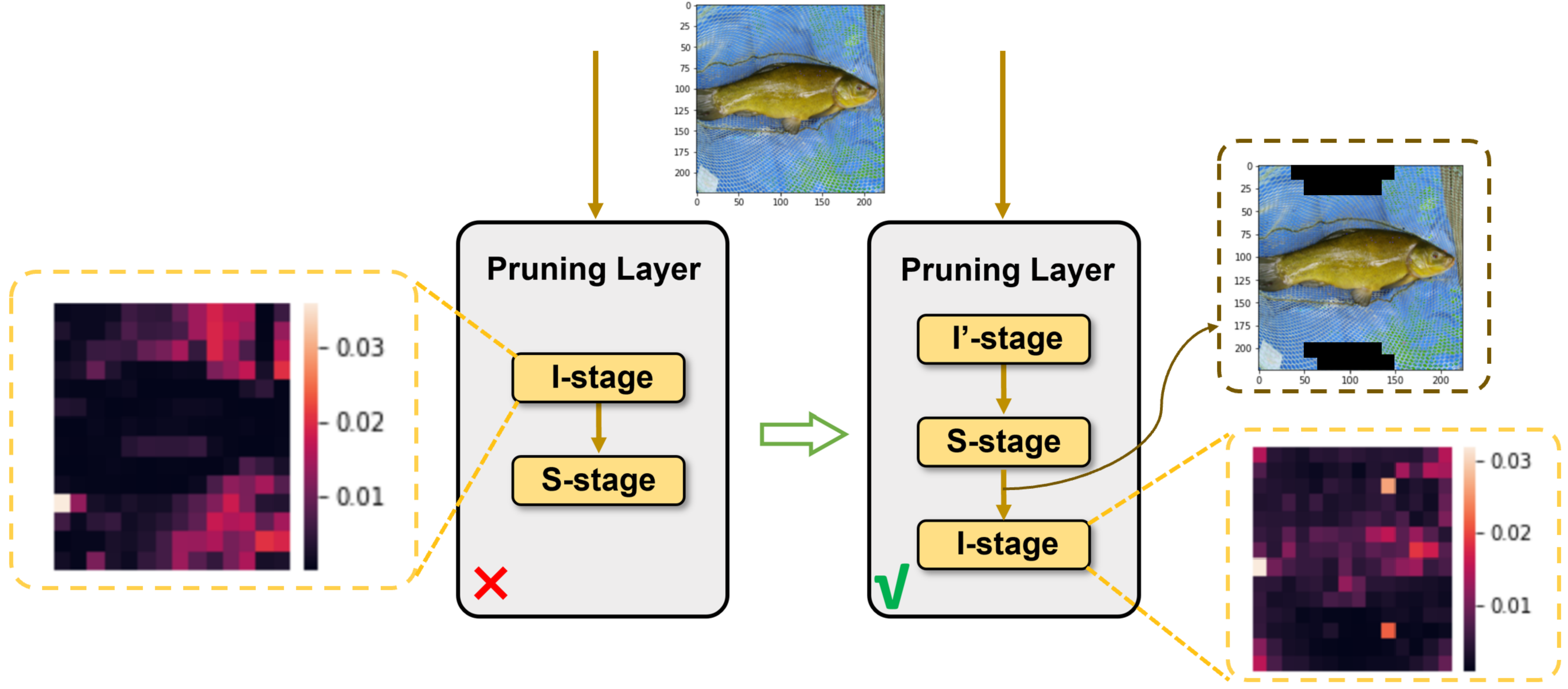}
\caption{Visual comparison between the I-S pattern and the I$^\prime$-S-I pattern.
}
\label{fig:solve}
\end{figure*}

As shown in Fig.~\ref{fig:solve}, by pruning similar background tokens in advance, the \textit{overwhelming of the major group} problem is alleviated significantly.

\section{Attention Probability Matrix}
\label{app:attention}

ViT \cite{vit} and its variants contain multiple Transformer encoder layers that are stacked up together. The basic Transformer 
encoder layer includes a multi-head attention (MHA) block followed by a feed-forward network (FFN) block, with residual 
connections and layer normalization around each. We make the assumption that an MHA block consists of $H$ independently 
parameterized heads. An attention head $h$ in layer $l$ can be parameterized by the Key, Query, and Value weight matrices: 
$\mathbf{W}_{k}^{(h, l)}, \mathbf{W}_{q}^{(h, l)}, \mathbf{W}_{v}^{(h, l)} \in \mathbb{R}^{d_{h} \times d}$, and the output 
weight matrix $ \mathbf{W}_{o}^{(h, l)} \in \mathbb{R}^{d \times d_{h}}$, where $d_{h}$ is typically set to $d/H$ and $d$ is 
the embedded feature dimension. Suppose $x \in \mathbb{R}^{d \times n}$ is the input sequence and $n$ is the input sequence 
length. For each head, the \emph{attention probability} between token $x_i$ and $x_j$ is given as an element of 
matrix $\mathbf{A}^{(h,l)}$:

\begin{equation}
    \mathbf{A}^{(h, l)}\left(x_{i}, x_{j}\right)=\operatorname{softmax}\left(\frac{x^{T} \mathbf{W}_{q}^{T} \mathbf{W}_{k} x}{\sqrt{d}}\right)_{(i, j)} \in \mathbb{R}
\end{equation}



This matrix measures how much token $x_j$ attends to token $x_i$. The output of an MHA block can be formulated as follows:

\begin{equation}
x_{\mathrm{MHA}}=\mathrm{LN}\left(\mathbf{W}_{o} \sum_{i=1}^{n} \mathbf{W}_{v} x_{i} \mathbf{A}^{(h, l)}\left(x_{i}, x_{j}\right)+x\right)
\end{equation}

The output of a Transformer encoder layer can be formulated as follows:

\begin{equation}
x_{\mathrm{out}}=\mathrm{LN}\left(\sigma\left(\mathbf{W}_{2}\left(\mathbf{W}_{1} x_{\mathrm{MHA}}+b_{1}\right)\right)+b_{2}+x_{\mathrm{MHA}}\right)
\end{equation}

\noindent
where $\mathbf{W}_{1}, \mathbf{W}_{2}, b_1$, and $b_2$ are FFN parameters, and $\sigma$ and $\mathrm{LN}$ denote the activation function and layer normalization, respectively. We can see that the computation overhead of a Transformer encoder layer undergoes a quadratic reduction when tokens are pruned.

\section{Optional Training Paradigm after Pruning}
\label{app:fine-tune}


Zero-TPrune can eliminate the fine-tuning process after pruning with a very small accuracy reduction. However, in some scenarios, we may have adequate samples and computational resources. In such cases, the performance of Zero-TPrune can be improved further by training (fine-tuning) after pruning. In this section, we introduce techniques used to accomplish this.

Given that it is very expensive to make importance-based ranking differentiable \cite{rankdiff}, we eliminate the 
\textbf{S-stage} and retain only the \textbf{I-stage} when we aim to further train (fine-tune) the pruned model.
Besides this, to make Zero-TPrune differentiable, it is necessary to replace normal token pruning with ``soft token pruning." Instead of completely discarding pruned tokens, soft token pruning assigns them small weights to reduce their effect on later computation and preserves compatibility with back-propagation during training. In this way, the non-differentiable token mask $M$ is replaced with a differentiable soft mask $\tilde{M}$ using the sigmoid operation:

\begin{equation}
\tilde{M}^{(l)}\left(x_i\right)=\sigma\left(\frac{s^{(l)}\left(x_i\right)-\theta^{(l)}}{T}\right)
\end{equation}

\noindent
where $s^{(l)}\left(x_i\right)$ is the importance score of token $x_i$ and $\theta^{(l)}$ is the importance threshold for 
the $l$-th layer. $\theta^{(l)}$ is determined based on the chosen pruning rate and GFLOPS budget. Details of soft token pruning 
can be found in \cite{learned}.


For simplicity, we use a similar loss function to DynamicViT \cite{dynamicvit}, which includes three terms:

\begin{equation}
    \mathcal{L}=\mathcal{L}_{\text {cls }}+\lambda_{\text {distill }} \mathcal{L}_{\text {distill }}+\lambda_{\mathrm{KL}} \mathcal{L}_{\mathrm{KL}}
\end{equation}

\noindent
The first term is the standard classification cross-entropy loss:

\begin{equation}
    \mathcal{L}_{\text {cls }}=\text { CrossEntropy }(\mathbf{y}, \overline{\mathbf{y}})
\end{equation}

\noindent
During fine-tuning, we use the original backbone network as the teacher model and push the behavior of the Zero-TPrune model to be as close to the teacher model as possible. First, we push the finally retained tokens of Zero-TPrune close to the ones of the teacher model. This contributes to the second distillation term above. We also minimize the difference in predictions between Zero-TPrune and its teacher via Kullback-Liebler (KL) divergence. This contributes to the third term. Details of the loss function can be found in \cite{dynamicvit}.

\section{Visualization}

In this section, we use some visualization examples to provide high-level insights.

\subsection{An Input Image Example}

\begin{figure*}[h]
\centering
\includegraphics[width=\linewidth]{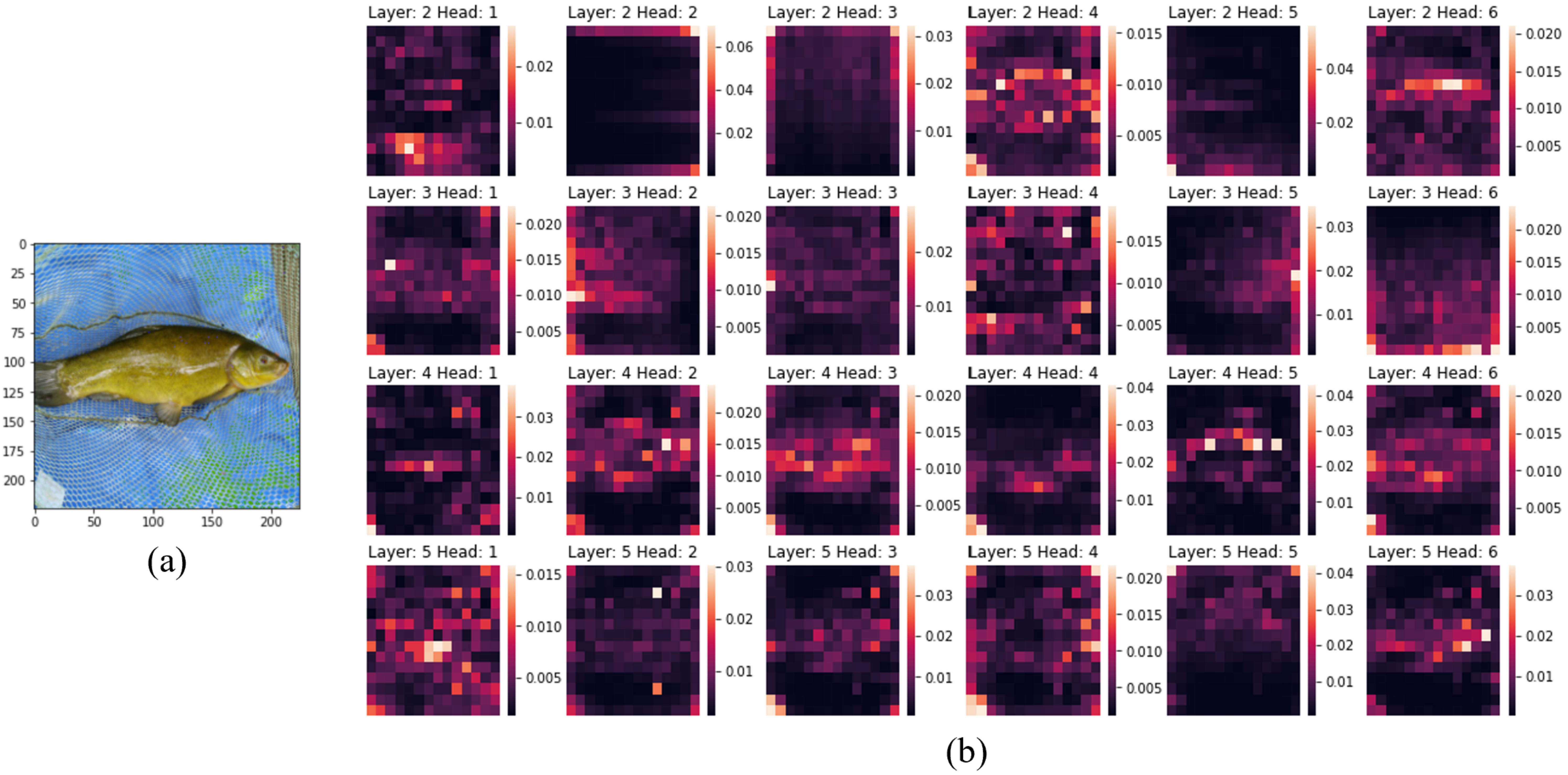}
\caption{The important part of input images can be successfully captured by the graph-based WPR algorithm: (a) a test
sample of \textit{fish} in the ImageNet dataset and (b) the corresponding importance score distributions given by the WPR algorithm in different layers. The used backbone is DeiT-S.}
\label{fig:vis1}
\end{figure*}

Fig.~\ref{fig:vis1} shows a simple test sample of a \textit{fish} from the ImageNet dataset and the corresponding importance score distributions in different layers and heads. We can see that most heads can successfully capture the important part of this image with the help of the graph-based WPR algorithm.

\subsection{Averaged Importance Distribution over Thousands of Images}

\begin{figure*}[h]
\centering
\includegraphics[width=\linewidth]{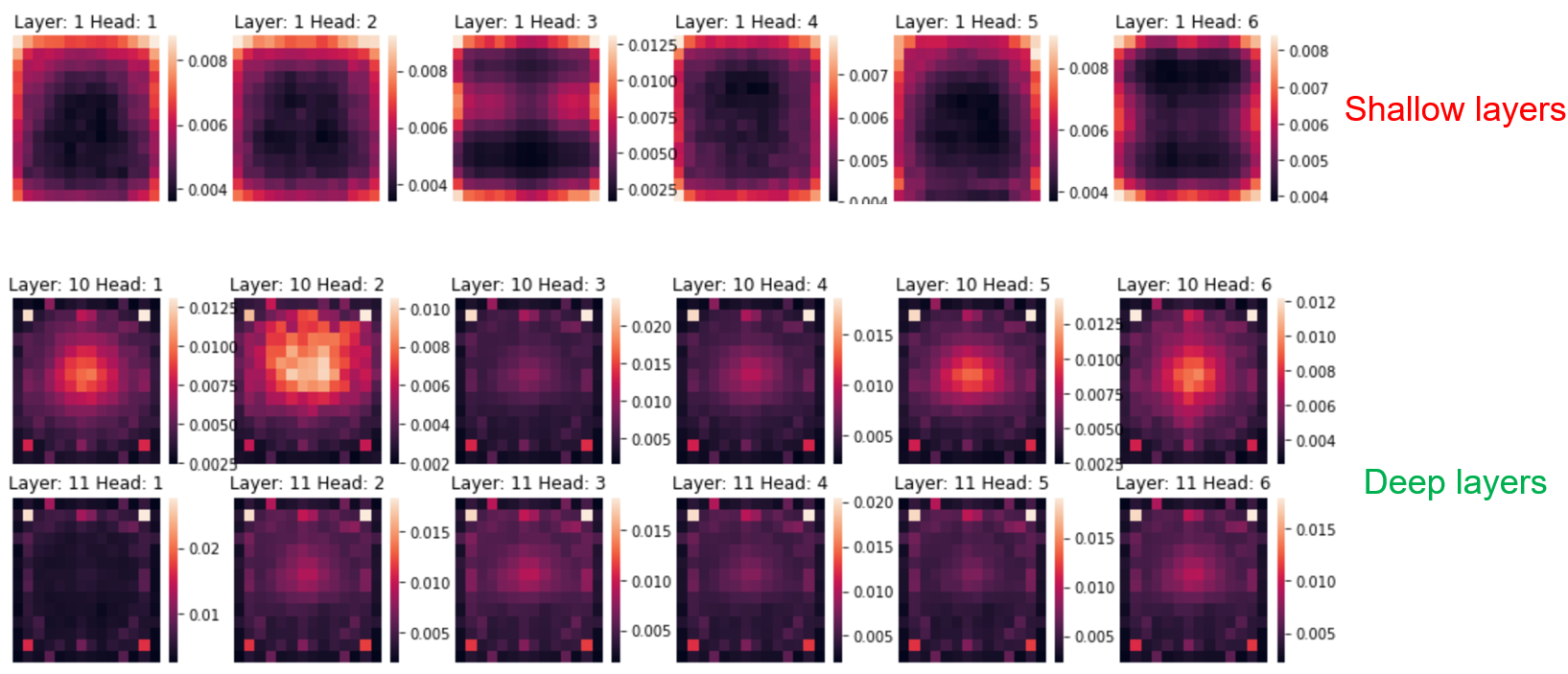}
\caption{Importance score distributions averaged over thousands of images. The first row is derived from the first layer and the second (third) row from the 10th (11th) layer of the DeiT-S model.}
\label{fig:vis2}
\end{figure*}

Another interesting visualization example is related to the general functionality of different layers in the Transformers. 
Fig.~\ref{fig:vis2} shows the importance score distributions averaged over thousands of images. It indicates that different layers of the Transformer behave differently. Shallow layers focus more on the edge of input images and deep layers focus more on the center.

\section{Combining Results of Different Heads}

In this section, we introduce the techniques we propose to nontrivially combine the importance score distribution of different heads from the WPR algorithm.

\subsection{Emphasizing Informative Region}
\label{app:eip}

\begin{figure*}[h]
\centering
\includegraphics[width=\linewidth]{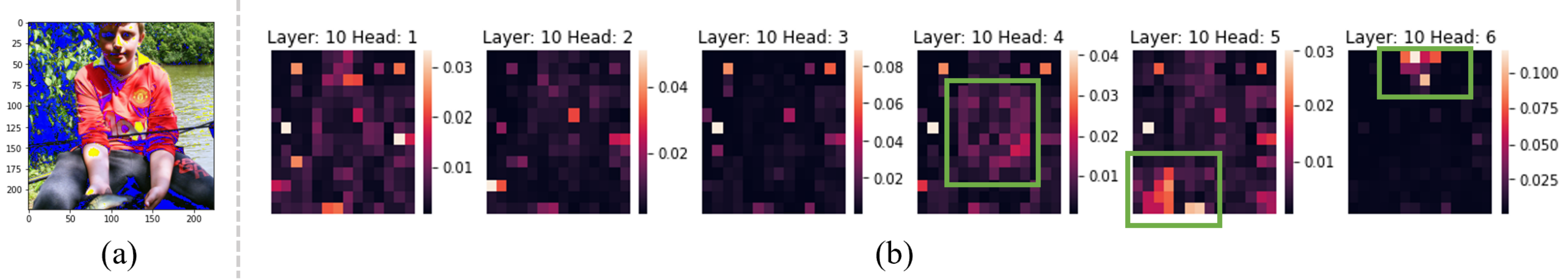}
\caption{The distribution of importance score from different heads for an input image: (a) an image of a boy holding a fish
and (b) importance score distributions. The results are obtained by the WPR algorithm with 30 iterations in the tenth layer of the DeiT-S model.}
\label{fig:diffhead}
\end{figure*}

Different heads in an encoder layer usually pay attention to different parts of the input image, as shown in Fig.~\ref{fig:diffhead}. For the input image of a boy holding a fish, some heads pay more attention to the body of this boy, some to the head of this boy, and some to the fish in hand. 

We propose EIR to address this issue. Suppose there are three heads in all and the importance scores of tokens $A$, $B$, and $C$ are [9,9,9], [9,0,0], [3,3,3], respectively. The ideal importance order is $A>B>C$. Table~\ref{tab:eip} shows the outcome of application of different importance score calculation methods. The traditional averaging method assigns the same importance to tokens $A$ and $B$. If we only select the highest score across all heads, tokens $A$ and $B$ will be assigned the same importance, which is also not desired. The proposed EIP technique balances the two situations and results in the ideal importance order.

\begin{table}[t]
  \small
  \caption{Application of different importance score calculation methods to the example.}
  \label{tab:eip}
  \centering
  \begin{tabular}{lccc}
    \toprule
    Importance Score     &  Average$\{S_i\}$     & $max\{S_i\}$ & EIP \\
    \midrule
    Token A & 9 & 9 & 5.2     \\
    Token B & 3 & 9 & 3     \\
    Token C & 3 & 3 & 1.7  \\
    Rank & $A>B=C$ & $A=B>C$ & \textbf{$A>B>C$} \\
    \bottomrule
  \end{tabular}
\end{table}

\subsection{Variance-based Head Filter}
\label{app:vhf}

\begin{figure}[h]
\centering
\includegraphics[width=\linewidth]{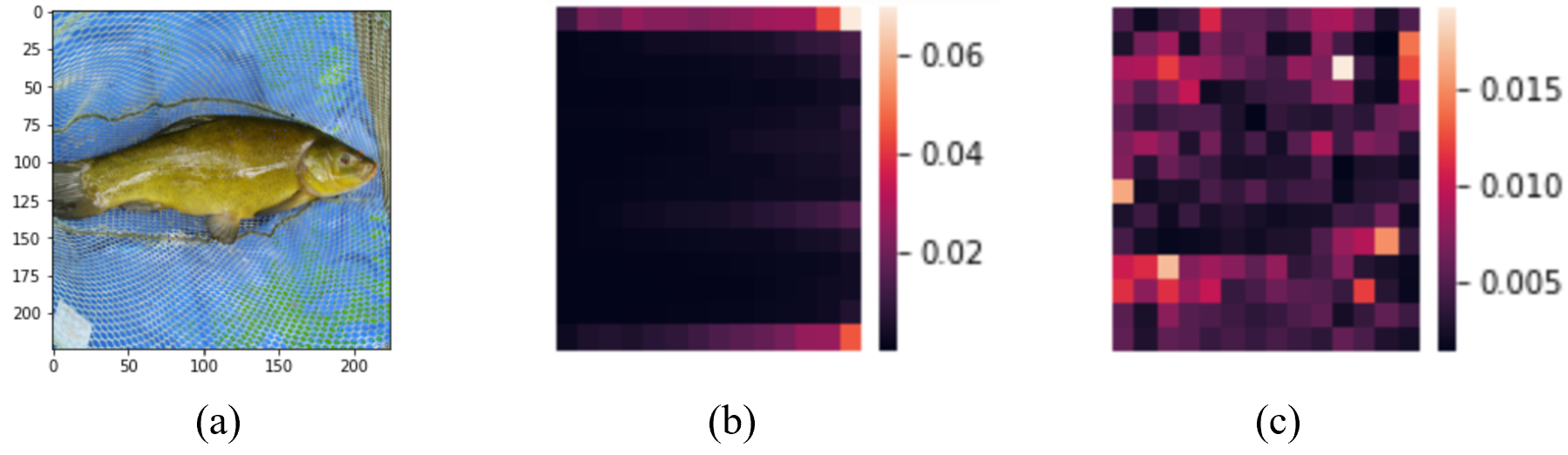}
\caption{Examples of undesired importance score distributions in certain heads obtained by the WPR algorithm: (a) input image, (b) second head in the second layer of the DeiT-S model, and (c) fourth head in the third layer of the DeiT-S model. }
\label{fig:vhf}
\end{figure}

The importance scores given by the WPR algorithm may converge to an undesired distribution. Two typical examples are shown in Fig.~\ref{fig:vhf}. Tokens at the edge of the input image get very high importance scores in Fig.~\ref{fig:vhf}(b) and the importance score distribution in Fig.~\ref{fig:vhf}(c) is nearly uniform. We introduce VHF to mitigate the negative impact of these heads.

\section{Downstream Tasks}
\label{app:downdata}

Table \ref{tab:downstream} shows the number of categories and test instances in the selected datasets. DTD is a 
describable textures dataset; Indoor67 is an indoor scene recognition dataset; CUB200 is a challenging dataset of 200 bird 
species. The other datasets have self-explanatory names.

\begin{table}[h]
  \caption{Datasets for downstream image classification.}
  \label{tab:downstream}
  \centering
  \begin{tabular}{lcc}
    \toprule
    Datasets     &  \#Categories     & \#Test Instances \\
    \midrule
    Flowers \cite{flower} & 102  & 6149     \\
    Pets \cite{pets}   & 37 & 3669      \\
    DTD  \cite{dtd}  & 47       & 1880  \\
    Indoor67 \cite{indoor}   & 67       & 1340  \\
    CUB200  \cite{cub}  & 200      & 5794  \\
    Aircrafts  \cite{aircrafts}  & 100       & 3333  \\
    Cars  \cite{cars}  & 196     & 8041  \\
    \bottomrule
  \end{tabular}
\end{table}

The experimental results are shown in Table~\ref{tab:transfer}. Zero-TPrune outperforms baselines on most datasets, indicating 
its strong transfer learning capability after pruning. ToMe has worse performance on small-sized models due to a lack of 
enough layers to merge tokens gradually.

\begin{table*}[t]
  \caption{Performance of pruned models on downstream tasks.}
  \label{tab:transfer}
  \centering
  \begin{tabular}{lcccccccc}
    \toprule
    Model & GFLOPS & Flowers & Pets  & DTD  &
    Indoor67 &  CUB200   & Aircrafts   &
    Cars \\
    \midrule
     Deit-T & 1.26  & 97.3 & 88.6 & 73.2 & 75.6 & 76.8 & 78.7 & 90.3     \\
     + ATS & 0.90 & 94.6 & 86.1 & \textbf{71.0} & 72.9 & 73.8 & 76.0 & \textbf{88.4}     \\
     + ToMe & 0.90 & 93.2 & 84.7 & 69.9 & 71.6 & 72.9 & 75.2 & 87.1  \\
     + Zero-TP & 0.91 & \textbf{95.1} & \textbf{86.9} & 70.9 & \textbf{73.7} & \textbf{74.4} & \textbf{76.7} &  88.2  \\
    \bottomrule
  \end{tabular}
\end{table*}

\section{Ablation Experiments}

\begin{figure*}[htbp]
\centering
\includegraphics[width=\linewidth]{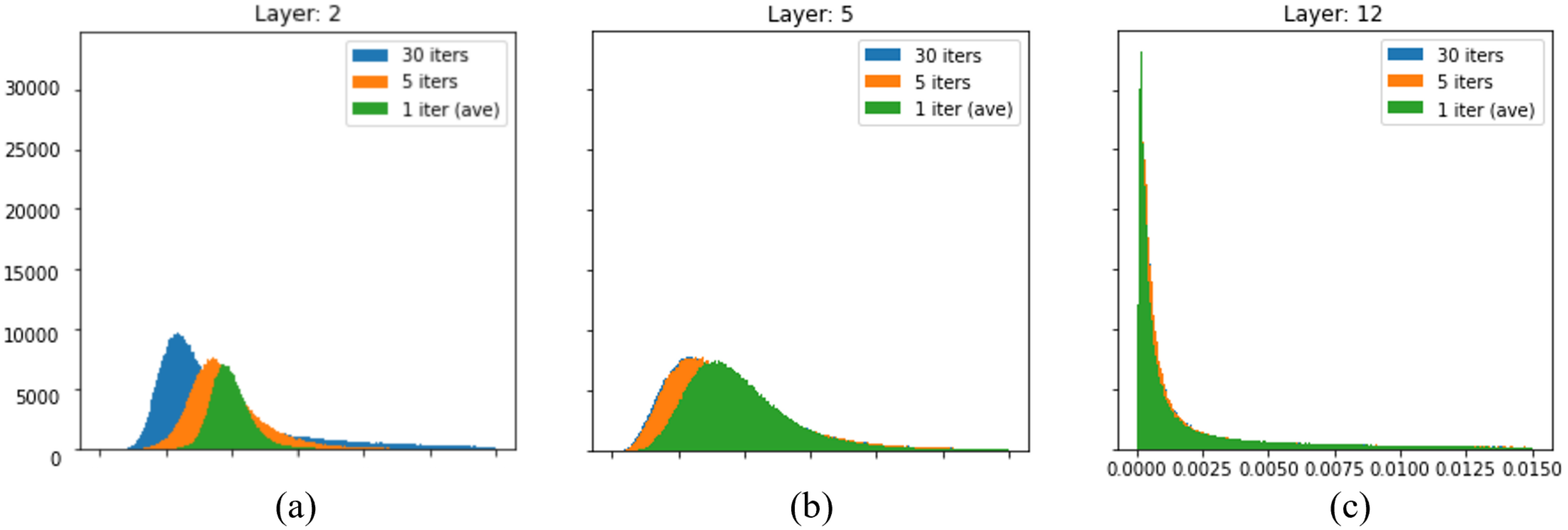}
\caption{The importance score distributions of tokens in 2,560 images. The distribution changes with both the number of iterations and layer location: (a) layer 2, (b) layer 5, and (c) layer 12 in DeiT-S.}
\label{fig:vis3}
\end{figure*}

In this section, we show results for further ablation experiments we performed. We explore the convergence
speed of WPR and determine the appropriate number of iterations for each layer in Section \ref{app:converge}. 
Then we identify a good enough variance threshold for VHF in Section \ref{app:range-vhf}. Furthermore, we 
describe optimal design choices in the \textbf{S-stage} in Section \ref{app:designchoice}, demonstrate the 
performance of Zero-TPrune-uni in Section \ref{app:zero-tp-uni}, and discuss hyperparameter search in Section \ref{app:hyper-search}.

\subsection{Convergence Speed of the WPR Algorithm}

\label{app:converge}

It is computationally expensive to check whether the WPR algorithm converges after each iteration. Thus, it would be desirable if we could determine the number of its iterations in advance. In order to do so, we need to derive the general convergence behavior of the WPR algorithm. Fig.~\ref{fig:vis3} shows the importance score distributions of tokens in 2,560 images. In the shallow layers, such as the first layer, the distributions corresponding to 30 iterations and five iterations are obviously different. This indicates that five iterations are not enough to make the WPR algorithm converge in the shallow layers. On the other hand, in the deep layers, such as the 12th layer, the distribution corresponding to 30 iterations is quite similar to the distribution corresponding to just one iteration. This means that one iteration is enough to make the WPR algorithm converge in the deep layers. In addition, in the fifth layer, five iterations are enough to make it converge. 

\begin{figure}[htbp]
\centering
\includegraphics[width=\linewidth]{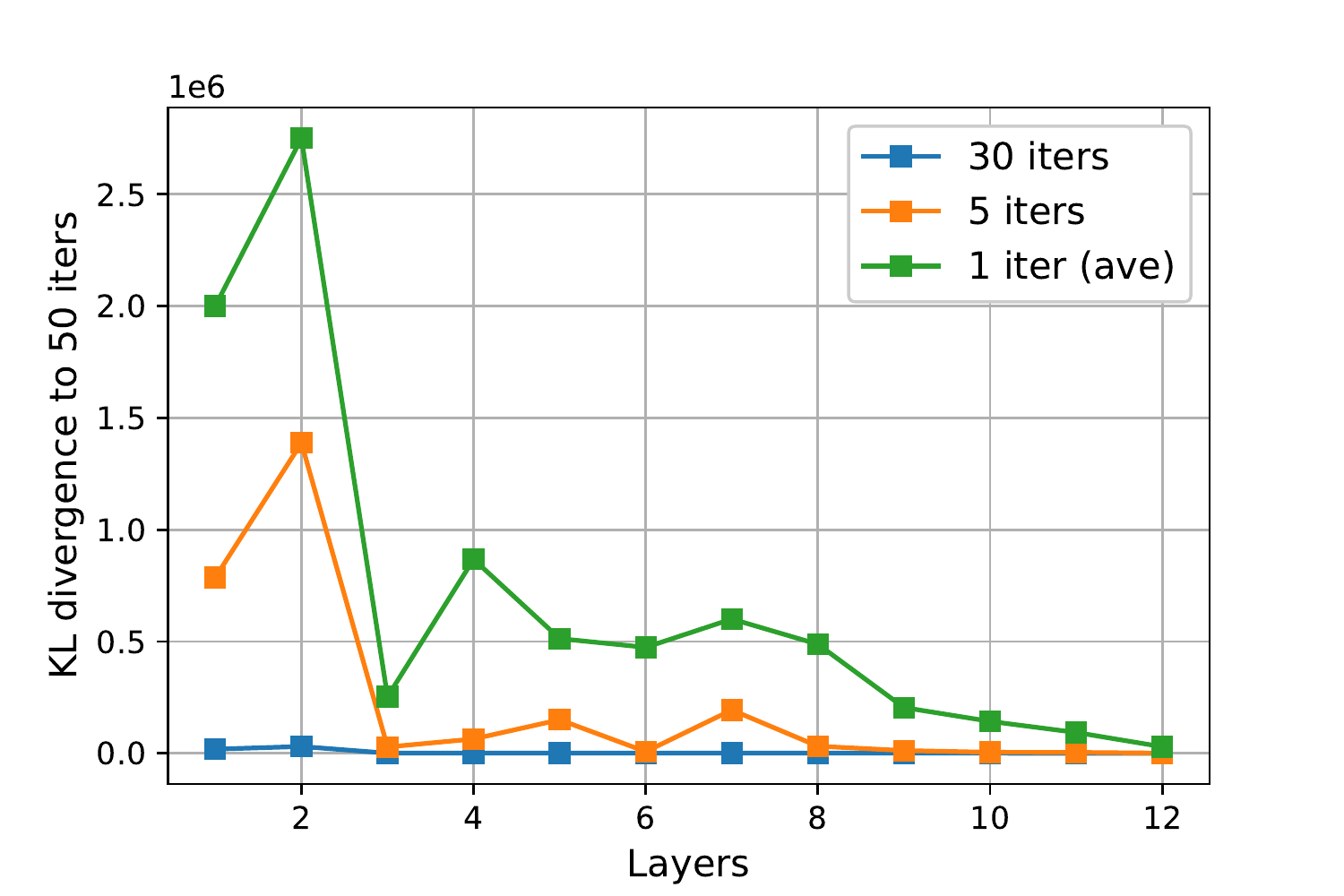}
\caption{The KL divergence between the importance score distribution given by different numbers of iterations and that given by 50 iterations in different layers. The used backbone is DeiT-S.}
\label{fig:kldiv}
\end{figure}

To quantitatively verify the assertions we made above, we calculate the KL divergence between the importance distribution
given by 30, 5, 1 iteration(s) and that given by 50 iterations in different layers. The results are shown in Fig.~\ref{fig:kldiv}. Thus, to ensure convergence, we set the number of iterations to 30-50, 5-10, and 1 in the first three layers, medium layers, and last three layers, respectively. Another interesting thing to note is that the Transformer model and the WPR algorithm assign low-importance scores to most tokens in the deep layers.

\subsection{Variance Thresholds for VHF}

\label{app:range-vhf}

\begin{figure}[h]
\centering
\includegraphics[width=\linewidth]{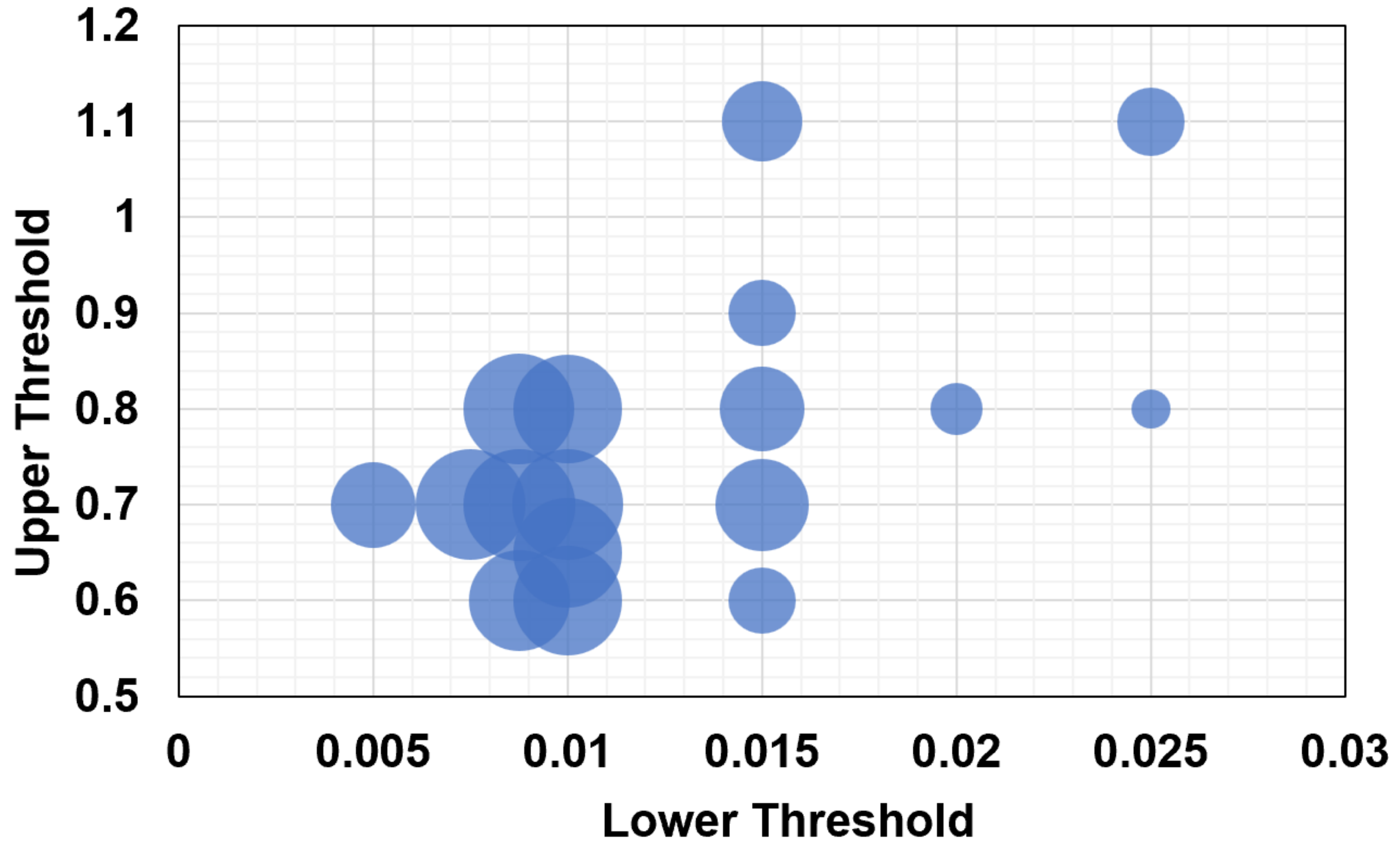}
\caption{Results obtained in the process of searching for optimal thresholds. A larger blue bubble represents higher 
accuracy with that setting.}
\label{fig:threshold}
\end{figure}

To exclude noise from heads that converge to undesired importance score distributions (as shown in Fig.~\ref{fig:vhf}), 
we propose VHF and set minimum and maximum thresholds for the variance of head distributions. We perform an ablation 
experiment to determine the optimal variance range. The pruning configuration is shown in Table \ref{tab:threshold}. We then 
use random initialization and beam search ($k=2$) to find a good enough variance range setting. The results are shown in 
Fig.~\ref{fig:threshold}, which points to the range [0.01,0.7].

\begin{table}[h]
  \caption{Pruning configuration used to search for optimal variance thresholds.}
  \label{tab:threshold}
  \centering
  \begin{tabular}{lcccccc}
    \toprule
    Pruning Layers     &  0 & 2 & 4 & 6 & 8 & 10 \\
    \midrule
    Retention Rates & 0.9 & 0.9 & 0.85 & 0.8 & 0.7 & 0.65  \\
    \# Iterations   & 50 & 50 & 5 & 5 & 1 & 1      \\
    \bottomrule
  \end{tabular}
\end{table}

\subsection{Optimal Design Choices in the S-stage}
\label{app:designchoice}

As discussed in Section \ref{s-stage}, the design space of the \textbf{S-stage} is composed of three dimensions: (1) source
of feature vectors, (2) partitioning method, and (3) similarity metric. We find that the optimal choice is (1) key matrix,
(2) sequential (prune unimportant part), and (3) cosine similarity, respectively. This is the default setting in the
following experiments unless otherwise noted. For the results in this section, pruning layers are inserted after the 
[1,3,6,9,11]-th layer with a retention rate of [1,0.9,0.8,0.7,1] and \#iterations of [30,5,5,1,1] in the \textbf{I-stage}, 
and 10 tokens are pruned in each \textbf{S-stage}. Note that all results in this subsection are augmented by the CLS token by assigning it an importance score that is $\sqrt{N}$ times larger than other tokens during initialization in the \textbf{I-stage}, where $N$ is the number of tokens.

\begin{figure}[t]
    \centering
    \includegraphics[width=\linewidth]{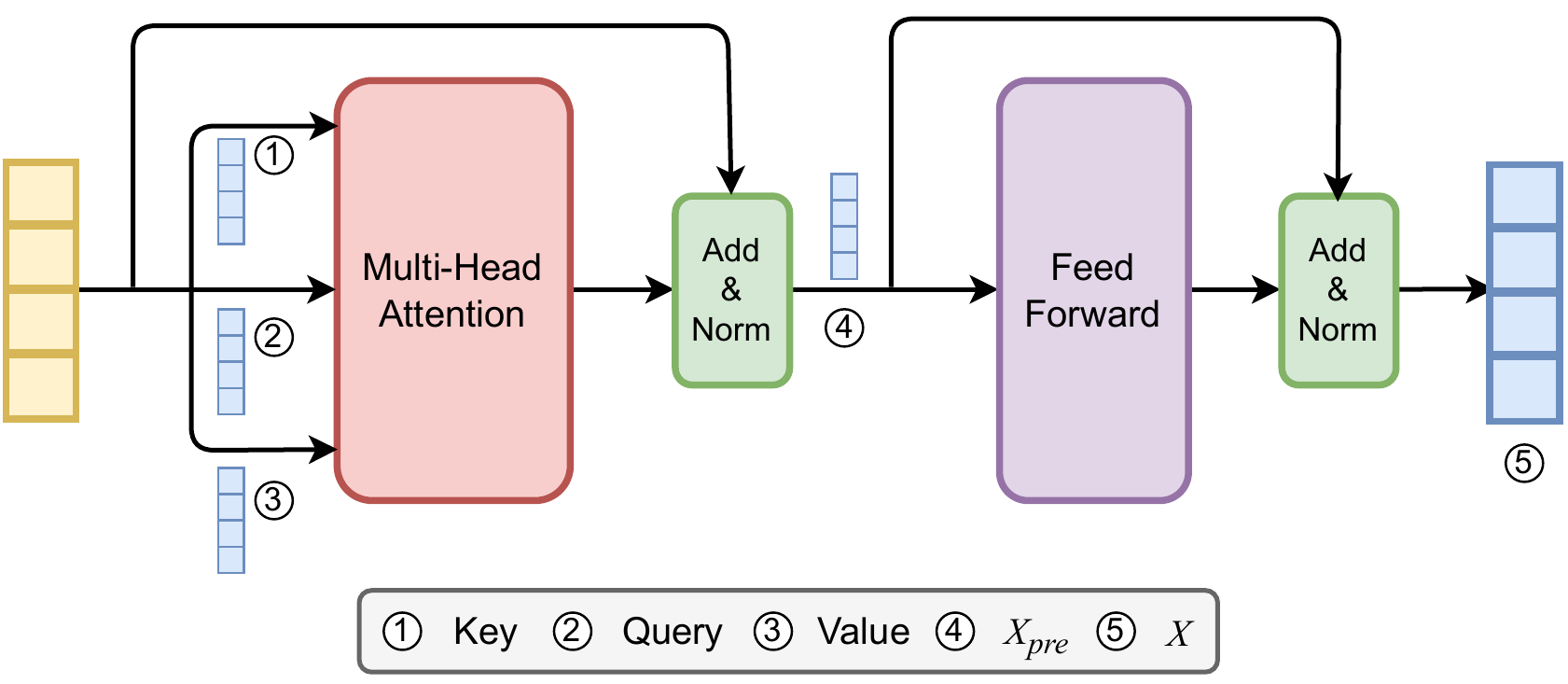}
    \captionof{figure}{Potential feature vectors that can be used to represent tokens.}
    \label{fig:feature}
\end{figure}

\textbf{Feature vectors:} As shown in Fig.~\ref{fig:feature}, feature vectors that represent tokens can be the corresponding vectors in the Key matrix, Query matrix, Value matrix, intermediate embedding vectors in the $X_{pre}$ matrix, or output embedding vectors in the $X$ matrix. We maintain the other settings and change the feature vectors used. The performance of pruned models is shown in Table~\ref{tab:feature}. It indicates that the Key matrix is the optimal source of feature vectors.

\begin{table}[t]
  \centering
  \caption{Ablation experiment results for the source of \textbf{feature vectors}.}
    \begin{tabular}{lcc}
      \toprule
        Feature  &  Acc@top1    & GFLOPS \\
        \midrule
        X$_{pre}$ & 79.113\% & 3.08     \\
        X   & 79.082\% & 3.08      \\
        \textbf{K} & \textbf{79.351\%}  & \textbf{3.08}  \\
        Q  &  79.205\%    & 3.08 \\
        V  & 79.097\%    & 3.08  \\
      \bottomrule
    \end{tabular}
  \label{tab:feature}%
\end{table}%


\textbf{Partitioning method:} After ranking tokens according to their importance (e.g., \textit{token
\{1,2,3,4,5,6\}}; \textit{token 1} has the highest score and \textit{token 6} has the lowest), we choose from the following options: (i) \textbf{Alternate}: alternatively assign them to Group $A$ and $B$, then the average token importance in two groups is nearly equal (e.g., $A:\{2,4,6\}$, $B:\{1,3,5\}$); (ii) \textbf{Sequential-U}: assign the less important half of tokens to Group $A$ and the other half to Group $B$, which means we sequentially partition tokens and prune the unimportant part (e.g., $A:\{4,5,6\}$, $B:\{1,2,3\}$); (iii) \textbf{Sequential-I}: assign the more important half of tokens to Group $A$ and the other half to Group $B$, which means we sequentially partition tokens and prune the important part (e.g., $A:\{1,2,3\}$, $B:\{4,5,6\}$), (iv) \textbf{Random}: randomly assign them to Group $A$ or $B$; and (v) \textbf{No partition}: assign all tokens to both groups without partitioning. To evaluate the effectiveness of these options, we conducted experiments while keeping all other settings at default values. The results are shown in Table \ref{tab:partition}, where Sequential-U represents choice (ii) and Sequential-I represents choice (iii). It clearly indicates that Sequential-U is preferable to all the other partitioning methods.

\textbf{Similarity metric:} We experimented with several metrics for measuring similarity between two vectors, including cosine similarity, dot product, and Minkowski distance with different $p$ values. When using Minkowski distance to measure similarity between vectors, we negated the distance to account for the fact that a longer distance indicates a lower similarity. The results of these experiments, shown in Table~\ref{tab:metrics}, indicate that cosine similarity is the best choice.

\begin{table}[htbp]
  \centering
    \caption{Ablation experiment results for choosing the \textbf{partitioning method}.}
      \label{tab:partition}
      \centering
      \begin{tabular}{lcc}
        \toprule
        Method   &  Acc@top1    & GFLOPS \\
        \midrule
        Random & 79.055\% & 3.08     \\
        Alternate   & 79.179\% & 3.08      \\
        \textbf{Sequential-U} & \textbf{79.351\%}  & \textbf{3.08}  \\
        Sequential-I   &  78.898\%    & 3.08 \\
        No partition  & 78.422\%    & 3.08  \\
        \bottomrule
      \end{tabular}
\end{table}%

\begin{table}[htbp]
  \centering
  \caption{Ablation experiment results for choosing the \textbf{similarity metric}.}
  \label{tab:metrics}
  \centering
  \begin{tabular}{lcc}
    \toprule
    Similarity  &  Acc@top1    & GFLOPS \\
    \midrule
    dot product & 79.257\% & 3.08     \\
    \textbf{cosine}   & \textbf{79.351\%} & \textbf{3.08}      \\
    Manhattan ($p=1$) & 79.208\%  &  3.07 \\
    Euclidean ($p=2$) & 79.224\%  &  3.07 \\
    Minkowski ($p=3$) & 79.246\%  &  3.07 \\
    Minkowski ($p=4$) & 79.273\%  &  3.07 \\
    Minkowski ($p=5$) & 79.189\%  &  3.07 \\
    Minkowski ($p=\infty$) & 79.092\%  &  3.07 \\
    \bottomrule
  \end{tabular}
\end{table}%

  
  

\subsection{Performance of Zero-TPrune-uni}
\label{app:zero-tp-uni}

The ablation experimental results of Zero-TPrune-uni are shown in Table \ref{tab:uni-ablation}. The backbone for deployment is DeiT-S, and the model is evaluated on the ImageNet validation set.

\begin{table*}[htbp]
  \centering
  \caption{Contribution breakdown of the different techniques employed in Zero-TPrune-uni. The ``-uni" suffix represents uniform initialization in the I-stage.}
  {
  \setlength\tabcolsep{3pt}
    \begin{tabular}{lllll}
    \toprule
    Acc@1  & Params & GFLOPS & Throughput (img/s) & Method \\
    \midrule
    79.8\% (base) & 22M & 4.55G &  1505.9 & Unpruned model \\
    76.8\% (-3.0\%) & 22M   & 3.08G  & 2164.4 & random drop \\
    78.0\% (+1.2\%) & 22M   & 3.08G & 2142.3 & WPR \\
    78.2\% (+0.2\%) & 22M   & 3.08G & 2139.6 & WPR + EIR \\
    78.4\% (+0.2\%) & 22M   & 3.08G & 2107.2 & WPR + EIR + VHF (I-stage) \\
    78.9\% (+0.5\%) & 22M & 3.08G & 2066.4 & I-stage + S-stage \\
    79.1\% (+0.2\%) & 22M & 3.08G & 2062.9 & I-stage + S-stage + MC Simulation \\
    
    \bottomrule
    \end{tabular}%
    }
  \label{tab:uni-ablation}%
\end{table*}%

\subsection{Hyperparameter Search}
\label{app:hyper-search}

\begin{figure}[t]
    \centering
    \includegraphics[width=\linewidth]{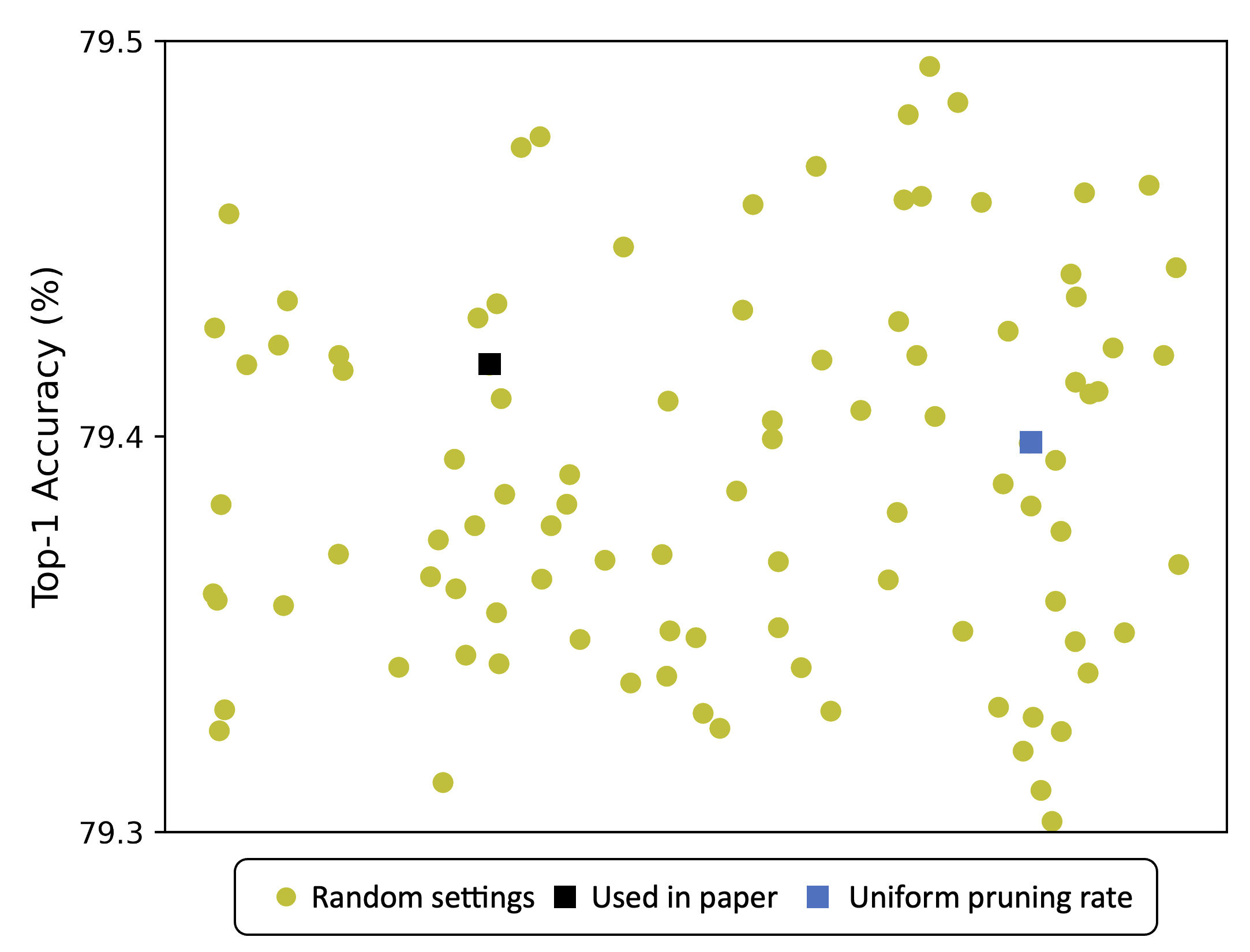}
    \captionof{figure}{One hundred randomly selected hyperparameter settings and their corresponding performance after 
being applied to DeiT-S without fine-tuning}
    \label{fig:hyper-search}
\end{figure}

The performance of Zero-TPrune, in terms of accuracy, is not sensitive to the hyperparameter setting as long as the 
number of pruning layers is more than two and the variance of their pruning rate is limited (i.e., the pruning process 
is not concentrated on one or two layers). We randomly choose different hyperparameter settings and show their 
performance in Fig.~\ref{fig:hyper-search}. This figure indicates that randomly selecting a hyperparameter setting 
does not hurt our performance much.

For a fair comparison with baselines, we do not use the best performance we can find through hyperparameter search. 
Instead, we use a hyperparameter setting with approximately the average performance among the search results. It is 
also close to setting a constant pruning rate across different layers. 

Even the full hyperparameter search process is much faster than fine-tuning. For hyperparameter search, we only need 
to perform inference. Specifically, for MCS, we randomly selected 1024 images in the validation dataset for each 
hyperparameter setting and obtain the corresponding accuracy. We tried 2000 settings on a single A100 GPU, which only 
required 3.8 hours. On the contrary, fine-tuning DeiT-S on the ImageNet dataset requires 144 A100 GPU hours.

\section{Comparison with State-of-the-Art Methods}

In this section, we first supplement comparisons with more depth-adaptive methods in Section \ref{app:depthada} 
and then compare Zero-TPrune with more straightforward attention-based token ranking methods in Section 
\ref{app:otherattn}. Finally, we provide performance comparisons with state-of-the-art fine-tuning-free token 
pruning methods in terms of throughput in Section \ref{app:throughput}.

\subsection{Depth-Adaptive Methods}
\label{app:depthada}

Token pruning can be seen as a fine-grained variant of the depth-adaptive transformer, such as layer dropping. One 
of our baselines, A-ViT \cite{avit}, is a token-wise depth-adaptive method. Instead of inserting pruning layers and setting pruning rates for them, it calculates the halting probability per token at each layer and halts tokens at adaptive depth. Zero-TPrune w/o fine-tuning competes with and even outperforms A-ViT w/ fine-tuning, as shown in Fig. \ref{fig:acc_flops}. A-ViT outperforms prior art on depth-adaptive methods. We adopt corresponding results and compare them with Zero-TPrune in Table \ref{tab:depthada}. Note that the result of Zero-TPrune is obtained off the shelf without fine-tuning, while other results are obtained after fine-tuning the adaptive models.

\begin{table}[t]
  \centering
  \caption{Comparison with depth-adaptive methods on the DeiT-T model. The performance of Zero-TPrune is obtained without fine-tuning, while other results are obtained with fine-tuning.}
    \begin{tabular}{lll}
    \toprule
    Method  &  Acc@top1    & GFLOPS \\
    \midrule
    DeiT-T \cite{deit} & 71.3\% & 1.3 \\
    \midrule
    ACT \cite{ACT} & 71.0\%  & 1.0 \\
    Confidence threshold \cite{confidencethreshold} & 65.8\%  & 1.1 \\
    Similarity gauging \cite{elbayad2019depth} & 69.4\%  & 1.1 \\
    PonderNet \cite{banino2021pondernet} & 66.2\% & 1.0 \\
    DynamicViT \cite{dynamicvit} & 70.9\% & 0.9 \\
    A-ViT \cite{avit} & 71.0\% & 0.8\\
    Zero-TPrune w/o FT &  70.4\%   & 0.9 \\
    \bottomrule
    \end{tabular}%
  \label{tab:depthada}%
\end{table}%

\subsection{Attention-based Token Ranking Methods}
\label{app:otherattn}

One of our baselines, ATS \cite{ats}, is an attention-based importance ranking method. It uses the attention
given by the CLS token to determine the importance of tokens. Simply averaging attention scores in the
attention matrix is the baseline of ATS (Fig. 3 in \cite{ats}) and performs worse than ATS. For the ablation
study, we replace our I-stage with top-$k$ importance selection based on (1) CLS token attention, (2)
average attention, and (3) accumulated average attention to improve the effectiveness of our method. Results
are shown in Table \ref{tab:attnbaseline}. The batch size is 512 and our \textbf{S-stage} is enabled in all settings. We adjust pruning rates slightly to match the FLOPs cost of different settings. Our proposed \textbf{I-stage} uses information from all tokens while reducing noise from 
unimportant tokens, leading to better performance.


\begin{table}[t]
  \footnotesize
  \centering
  \caption{Performance of pruned DeiT-S models without fine-tuning. Throughput is measured on a single NVIDIA A100 GPU.}
    \begin{tabular}{llll}
    \toprule
    Method  &  Acc@top1    & GFLOPS & Throughput(img/s) \\
    \midrule
    DeiT-S & 79.8\% & 4.55 & 1505.9    \\
    CLS Attn.   & 78.9\%  & 3.00  & 2179.3  \\
    Ave. Attn. & 78.4\%  & 2.99 & 2185.2 \\
    Accu. Ave. Attn.   & 78.5\%  & 2.97 & 2189.2 \\
    \textbf{I-stage}  &  \textbf{79.4\%}   & 2.97  & 2188.4  \\
    \bottomrule
    \end{tabular}%
  \label{tab:attnbaseline}%
\end{table}%

\subsection{Fine-Tuning-Free Token Pruning Methods}
\label{app:throughput}

\begin{figure}[t]
    \centering
    \includegraphics[width=\linewidth]{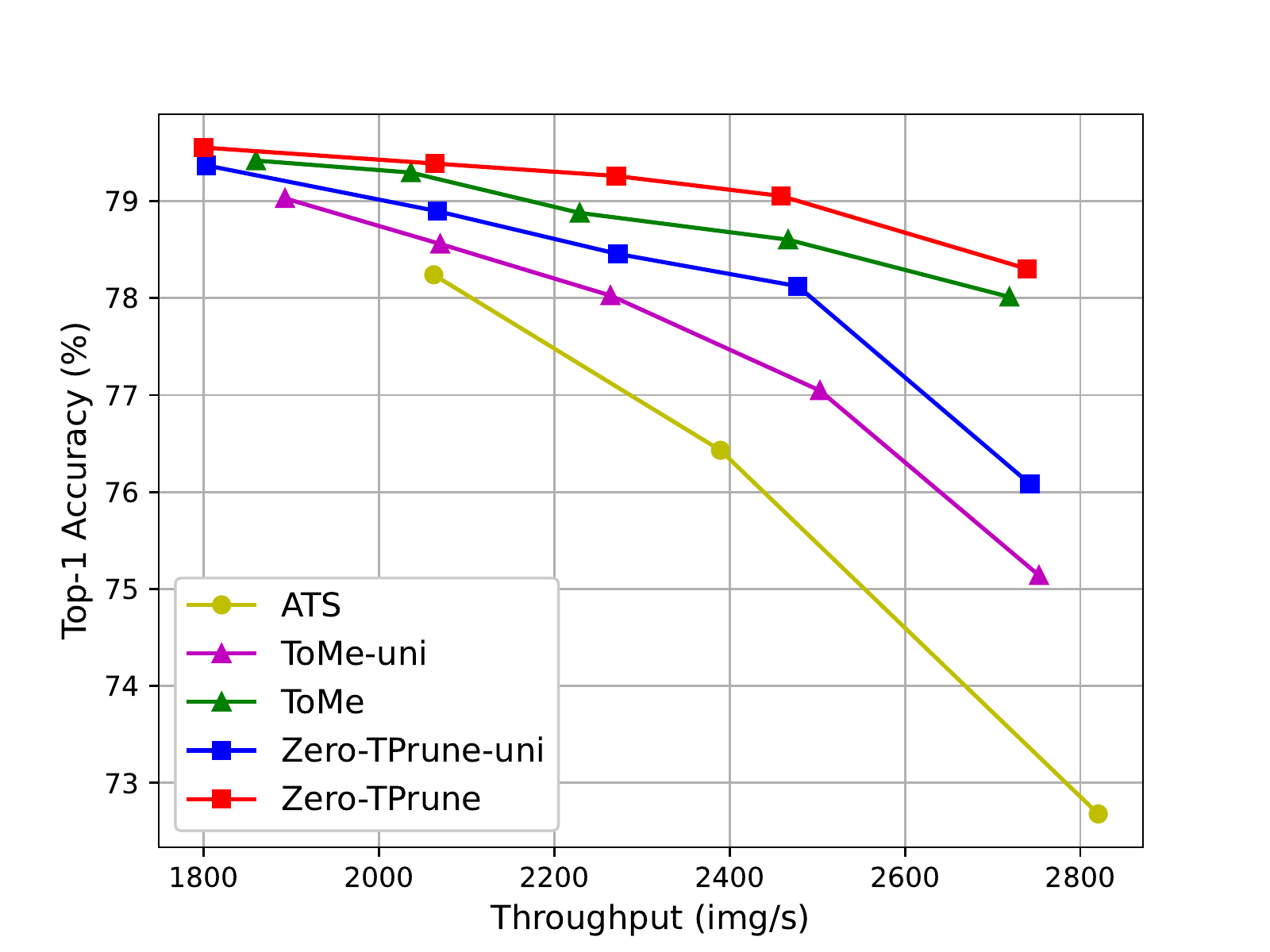}
    \caption{Performance comparison between Zero-TPrune and state-of-the-art fine-tuning-free methods. The applied Transformer backbone is DeiT-S.}
    \label{fig:accu_throughput}
\end{figure}

We conduct experiments on DeiT-S to show the superiority of Zero-TPrune over state-of-the-art fine-tuning-free token pruning/merging methods. The experimental results showing the trade-off between accuracy and throughput are shown in Fig. \ref{fig:accu_throughput}.

\section{Comparison between Scaling and Pruning}

\label{app:scaling}

\begin{figure*}[htbp]
\centering
\includegraphics[width=\linewidth]{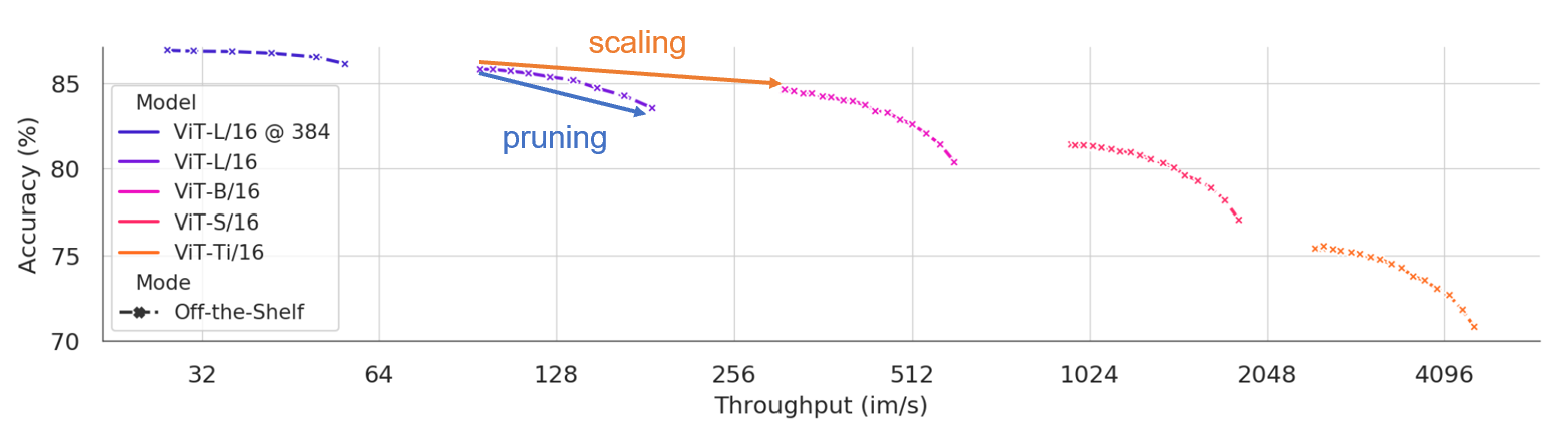}
\caption{Off-the-shelf performance of ViT models under ToMe \cite{tome}. This figure is adopted from \cite{tome}.}
\label{fig:ToMe}
\end{figure*}

As shown in Table \ref{tab:augreg}, Zero-TPrune cannot outperform all baseline methods when a relatively high pruning rate (e.g., reduce GFLOPS by 50\%) is applied to large models (e.g., DeiT-L). However, in this case,  scaling to a smaller model is often a better choice. ToMe outperforms Zero-TPrune when large models are aggressively pruned. Thus, we use the results from ToMe to illustrate this point.

\begin{table}[t]
  \centering
  \caption{Performance of pruned AugReg, LV-ViT, and SWAG models without fine-tuning. SWAG models perform inference on 384px images.}
    \begin{tabular}{rrll}
    \toprule
    Method  &  Acc@top1    & GFLOPS \\
    \midrule
    LV-ViT-M & 84.0\% & 12.7     \\
    + ATS   & 80.9\% & 6.4     \\
    + ToMe & \textbf{81.6\%} & \textbf{6.3}\\
    + Zero-TP   &  81.4\%   & 6.3 \\
    \midrule
    MAE & 83.62\% & 55.4     \\
    +ATS & 78.39\%  &  29.1 \\
    +ToMe & \textbf{78.95\%}  &  28.8 \\
    +Zero-TP   & 78.94\% & 28.6     \\
    \midrule
    SWAG & 85.30\% & 55.6     \\
    +ATS & 81.03\%  &  27.8 \\
    +ToMe & \textbf{84.59\%}  &  28.4 \\
    +Zero-TP   & 84.04\% & 28.3      \\
    \bottomrule
    \end{tabular}%
  \label{tab:augreg}%
\end{table}%

In Fig.~\ref{fig:ToMe}, ToMe is applied to different ViT backbones with different configurations. Different points on the same 
curve represent different configurations applied to the same backbone. The first point from the left on each curve represents 
the unpruned model. Aggressively pruning a model implies switching from the first point from the left on a given curve to the 
last point on this curve, which increases throughput but suffers from lower accuracy. Switching from the first point from the 
left on a given curve to the first point on another curve directly scales the size of the model without pruning. 
Aggressively pruning large models (ViT-L and ViT-B) underperforms scaling them in terms of both 
accuracy and throughput. On the contrary, for the ViT-S model, although scaling outperforms aggressive pruning in terms of 
throughput, it achieves lower accuracy than aggressive pruning.


\end{document}